\documentclass{article}

\usepackage{microtype}
\usepackage{graphicx}
\usepackage{subfig}
\usepackage{booktabs} % for professional tables

\usepackage{hyperref}

\usepackage[accepted]{icml2025}

\usepackage{amsmath}
\usepackage{amssymb}
\usepackage{mathtools}
\usepackage{amsthm}

\usepackage{enumitem}
\usepackage[capitalize,noabbrev]{cleveref}

\theoremstyle{plain}

\theoremstyle{definition}

\theoremstyle{remark}

\usepackage[textsize=tiny]{todonotes}

\usepackage{comment}
\usepackage{mathdots}
\usepackage{multirow}
\usepackage{lscape}
\usepackage{tabularx}
\usepackage{diagbox} % For diagonal lines in table cells

\newcommand{\vect}[1]{\boldsymbol{#1}}
\usepackage{longtable}

\usepackage{pifont}% http://ctan.org/pkg/pifont
\newcommand{\xmark}{\ding{55}}%

\newcommand{\blue}[1]{\textcolor{blue}{#1}}

\icmltitlerunning{Concept-Based Explainable Artificial Intelligence:
Metrics and Benchmarks}

\begin{document}

\twocolumn[
\icmltitle{Concept-Based Explainable Artificial Intelligence:
Metrics and Benchmarks}



\icmlsetsymbol{equal}{*}

\begin{icmlauthorlist}
\icmlauthor{Halil Ibrahim Aysel}{yyy}
\icmlauthor{Xiaohao Cai}{yyy}
\icmlauthor{Adam Prugel-Bennett}{yyy}
%\icmlauthor{Firstname4 Lastname4}{sch}
%\icmlauthor{Firstname5 Lastname5}{yyy}
%\icmlauthor{Firstname6 Lastname6}{sch,yyy,comp}
%\icmlauthor{Firstname7 Lastname7}{comp}
%\icmlauthor{Firstname8 Lastname8}{sch}
%\icmlauthor{Firstname8 Lastname8}{yyy,comp}
\end{icmlauthorlist}

\icmlaffiliation{yyy}{Electronics and Computer Science, University of Southampton, Southampton, United Kingdom}
%\icmlaffiliation{comp}{Company Name, Location, Country}
%\icmlaffiliation{sch}{School of ZZZ, Institute of WWW, Location, Country}

\icmlcorrespondingauthor{Halil Ibrahim Aysel}{hia1v20@soton.ac.uk}
%\icmlcorrespondingauthor{Firstname2 Lastname2}{first2.last2@www.uk}

\icmlkeywords{Machine Learning, ICML}

\vskip 0.3in
]



\printAffiliationsAndNotice{}  % leave blank if no need to mention equal contribution

\begin{abstract}
Concept-based explanation methods, such as concept bottleneck models (CBMs), aim to improve the interpretability of machine learning models by linking their decisions to human-understandable concepts, under the critical assumption that such concepts can be accurately attributed to the network's feature space. However, this foundational assumption has not been rigorously validated, mainly because the field lacks standardised metrics and benchmarks to assess the existence and spatial alignment of such concepts. To address this, we propose three metrics: the \textit{concept global importance metric}, the \textit{concept existence metric}, and the \textit{concept location metric}, including a technique for visualising concept activations, i.e., \textit{concept activation mapping}. We benchmark post-hoc CBMs to illustrate their capabilities and challenges. Through qualitative and quantitative experiments, we demonstrate that, in many cases, even the most important concepts determined by post-hoc CBMs are not present in input images; moreover, when they are present, their saliency maps fail to align with the expected regions by either activating across an entire object or misidentifying relevant concept-specific regions. 
We analyse the root causes of these limitations, such as the natural correlation of concepts. Our findings underscore the need for more careful application of concept-based explanation techniques especially in settings where spatial interpretability is critical.
\end{abstract}

\section{Introduction}
\label{sec:intro}

In recent years interest in explainable artificial intelligence (XAI) methods has grown substantially because of the desire to exploit the success of newly developed machine learning methods to new areas of our lives \cite{goebel2018explainable,  buhrmester2021analysis, van2022explainable, ali2023explainable, hassija2024interpreting}. In an attempt to make XAI more understandable to the layman there has been a growing drive to develop techniques that provide explanations in terms of human-understandable concepts \cite{bau2017network, kim2018interpretability, koh2020concept, havasi2022addressing, aysel2023multilevel, shin2023closer}. One of the big challenges of concept-based XAI methods is that of paramount importance yet lacks a systematic study to ensure that the concepts identified as important to making a decision properly align with human understanding of the concepts.  

In this paper, we propose three new metrics for measuring this alignment. The first is the \textit{concept global importance metric} (CGIM), measuring the concept alignment  for each image in a class. The second is the \textit{concept existence metric} (CEM), measuring whether the concepts identified as important for making a classification exist in an image. For example, if the horn is identified as the most important concept for deciding the image is a rhinoceros then we should expect the horn to be visible in the image. The third metric is the \textit{concept location metric} (CLM), measuring whether the excitable region of the feature maps used to determine an important concept is close to the location where we would expect the concept to be. In the example above, we would expect the heatmap representing the area of the feature map that corresponds to the horn concept should be located around the horn. Using these three metrics, we create a benchmark problem using the \texttt{Caltech-UCSB Bird (CUB)} dataset \cite{wah2011caltech}, and test the performance of concept-based XAI methods. 

%This is a rich dataset that provides 200 bird classes together with 112 binary concepts and the locations of many parts of the bird given for each image. Using this dataset and our new metrics we can test the performance of concept-based XAI methods over the whole dataset of 11,800 images.

%%%%%%newly added Figure
\begin{figure*}[!htb]
    \centering    \includegraphics[width=\linewidth]{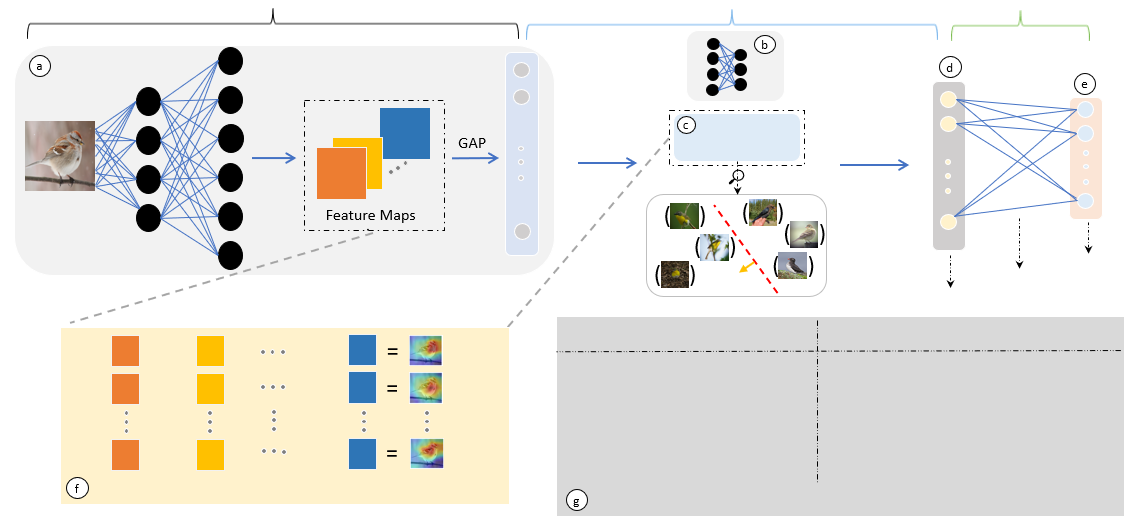}
    \put(-470, 132){$\vect{X}_i$}
    \put(-390, 226){$f:\mathcal{X} \rightarrow \mathbb{R}^d$}
    \put(-203, 226){$g:\mathbb{R}^d \rightarrow \mathbb{R}^{L}$}
    \put(-77, 226){$h:\mathbb{R}^{L} \rightarrow \cal{Y}$}
    \put(-191,167){$\vect{C} \in \mathbb{R}^{L \times d}$}
    \put(-197,160){\scriptsize Concept Vectors}
    \put(-188,104){\scriptsize $\vect{c}_{\rm breast}$}
    \put(-210,131){\small $f$}
    \put(-197, 118){\small $f$}
    \put(-214, 106){\small $f$}
    \put(-177, 132){\small $f$}
    \put(-159, 122){\small $f$}
    \put(-163, 110){\small $f$}
    \put(-457, 75){\scriptsize $c_{11} *$}
    \put(-457, 58){\scriptsize $c_{21} *$}
    \put(-458, 30){\scriptsize $c_{L1} *$}
    \put(-427, 75){\scriptsize $+ c_{12} *$}
    \put(-427, 58){\scriptsize $+ c_{22} *$}
    \put(-427, 30){\scriptsize $+ c_{L2} *$}
    \put(-390, 75){\scriptsize $+$}
    \put(-390, 58){\scriptsize $+$}
    \put(-390, 30){\scriptsize $+$}
    \put(-365, 75){\scriptsize $+ c_{1d} *$}
    \put(-365, 58){\scriptsize $+ c_{2d} *$}
    \put(-365, 30){\scriptsize $+c_{Ld} *$}
    \put(-294, 75){\scriptsize $\vect{F}_{i1}$}
    \put(-294, 58){\scriptsize $\vect{F}_{i2}$}
    \put(-294, 30){\scriptsize $\vect{F}_{iL}$}
    \put(-400, 13){\scriptsize \it \blue{CoAM Framework}}
    \put(-109, 95)  {\scriptsize $\hat{\vect{u}}_i = g(f(\vect{X}_i))$}
    \put(-56, 103){\scriptsize $\vect{\theta}$}
    \put(-28, 108){\scriptsize $\hat{\vect{y}}_i$}
    \put(-225, 80){\scriptsize \textbf{Global Evaluation}}
    \put(-250, 65){\scriptsize \textbullet\ Global concept importance:  $\vect{\theta}$}
    \put(-250, 55){\scriptsize \textbullet\ Concept vector $j$: $\vect{\theta}(j,:)$}
    \put(-250, 45){\scriptsize \textbullet\ Ground truth for concept $j$: $\vect{V}(j,:)$}
    \put(-250, 35){\scriptsize \textbullet\ CGIM: calculate similarity between}
    \put(-247, 25){\scriptsize $\vect{\theta}(j,:)$ and $\vect{V}(j,:)$}
     \put(-108, 80){\scriptsize \textbf{Local Evaluation}}
    \put(-135, 65){\scriptsize \textbullet\ Local concept importance:  $\theta_{jk} * \hat{u}_{ij}$}
    \put(-135, 55){\scriptsize \textbullet\ Choose top $l$ most important concepts}
    \put(-135, 45){\scriptsize \textbullet\ CEM: check if these concepts exist}
    \put(-135, 35){\scriptsize \textbullet\ Create concept-wise heatmaps via \textcircled{f}}
    \put(-135, 25){\scriptsize \textbullet\ CLM: check if each heatmap contains the}
    \put(-128,17){\scriptsize centre pixel location of each concept}
    \put(-200, 8){\scriptsize \it \blue{Proposed metrics: CGIM, CEM and CLM}}
\caption{\textbf{Overview of CAVs, CBMs, post-hoc CBMs and the proposed techniques.} Feature extractor \textcircled{a}, concept prediction block \textcircled{b}, CAVs \textcircled{c}, concept bottleneck \textcircled{d}, classifier \textcircled{e}, and our proposed CoAM framework \textcircled{f}. A traditional (without concept bottleneck) classification model consists of \textcircled{a} + \textcircled{e}, and \textcircled{c} is the introduced post-hoc to explain its predictions via CAVs. \textcircled{a} + \textcircled{b} + \textcircled{d} + \textcircled{e} forms the steps for traditional CBMs training, whereas \textcircled{a} + \textcircled{c} + \textcircled{d} + \textcircled{e} forms the post-hoc CBMs. Our proposed CoAM framework is \textcircled{f}, weighing pre-GAP feature maps with CAVs for concept visualisation. \textcircled{g} presents the example steps of our proposed metrics.}
    \label{fig:pcbms}
\vspace{-0.08in}
\end{figure*}
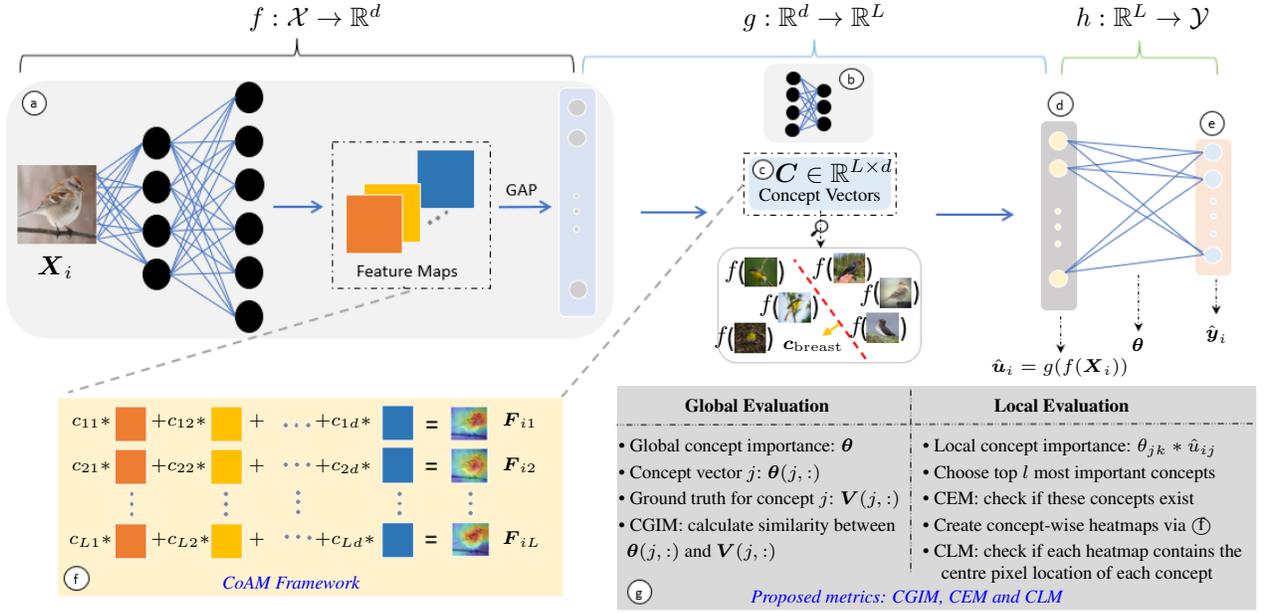

To illustrate the usefulness of our metrics, we examine a prominent example of a concept-based XAI system known as the post-hoc concept-bottleneck models (CBMs) \cite{yuksekgonul2023posthoc}. This method is designed to provide explanations of classifiers based on deep neural network (DNN). The method is a synthesis of two approaches -- traditional CBMs \cite{koh2020concept} and concept activation vectors (CAVs) \cite{kim2018interpretability} -- for concept-based explanations. Traditional CBMs are a relatively straightforward approach to introducing human-understandable concepts into XAI. In traditional CBMs, we start from a network trained to classify a set of classes and replace the final few layers with a new set of layers that are trained to predict human-understandable concepts, which provides a “concept bottleneck”.  From this concept representation, a fully connected layer is trained to predict the classes. Given a new image, it is then straightforward to see which concepts are important in making the prediction \cite{koh2020concept}. The disadvantage of traditional CBMs is that in order to train the network it requires that every image is annotated with the set of concepts that are visible in the image. Although there exists a few datasets where such annotations are given, generally it would be prohibitively expensive to annotate a large dataset.  

There has therefore been a drive to find cheaper methods to learn concepts.  One example proposed by Aysel et al. \cite{aysel2023multilevel} is to use annotations for the classes rather than individual images.  A second family of models that were developed to provide concept-based explanations is known as CAV methods \cite{kim2018interpretability}. These methods take a pre-trained network and probe the internal representation to determine the directions in that representation that align with human-understandable concepts.  One approach for doing this is to take two subsets of the images, one class where the concept is present and the other class where the concept is absent.  From examining the difference in the representations between the two classes we can determine CAVs. This method is an example of a “post-hoc” XAI method as it seeks to explain the decisions of a pretrained network without changing that network.

The post-hoc CBMs \cite{yuksekgonul2023posthoc} combine CAVs with the traditional CBMs. They take a pre-trained network and feed each channel in the last convolution layers into a global average pooling (GAP) layer. They use the GAP representation to learn a set of CAVs.  To do so, for each concept, they choose $m$ positive and negative example images which they then train a support vector machine (SVM) to separate ($m$ is of the order of 100 images). Each SVM discriminant vector is taken as a CAV. These CAVs are then used to determine the degree to which a concept is present in an image. From this, a concept bottleneck can be trained. This is the post-hoc CBMs that we study in this paper. The network is illustrated in the top row of Figure \ref{fig:pcbms}, and the bottom row illustrates the new metrics that we propose to evaluate the alignment of the concepts with human understanding of the concepts.

Although post-hoc CBMs sacrifice some performance accuracy in predicting classes compared to traditional DNNs (i.e., the ones without a concept bottleneck), they provide a relatively cheap way to obtain human-interpretable concepts. However, for this explanation to be useful, the concepts need to be accurately aligned with human understanding of the concepts. We use the new metrics and new benchmarks to evaluate this alignment. As we will see, the alignment is surprisingly poor, which highlights the necessity of introducing new metrics for assessing this alignment. The main contributions of the paper are as follows.
\vspace{-0.12in}
\begin{itemize}  [noitemsep]
    \item We propose the {\it concept activation mapping} (CoAM) to visualise concept activations.  
    \item We propose three quantitative metrics: i) CGIM, to test the global concept alignment by XAI methods; ii) CEM, to test whether a concept being identified by XAI methods exists in the image; and iii) CLM, to test whether a concept being identified by XAI methods is spatially aligned with the human concept.
    \item We benchmark the post-hoc CBMs \cite{yuksekgonul2023posthoc} using the proposed metrics to evaluate the alignment of concept-based XAI techniques on a benchmark dataset and discuss their advantages and limitations. 

\end{itemize}

% The rest of the paper is structured as follows. Section \ref{sec:related} recalls related work on concept-based explainability. Section \ref{sec:preliminary} provides an overview of CBMs and CAVs as foundational methodologies. Section \ref{sec:methodology}  introduces our proposed metrics CGIM, CEM, and CLM, including our CoAM framework for concept activation visualisation. The thorough experimental results are presented in Section \ref{sec:experiments}, followed by a detailed discussion of the findings in Section \ref{sec:discussion} and limitations. Finally, Section \ref{sec:conclusion} concludes the paper.

\section{Preliminary}
\label{sec:preliminary}

%In this section, we first present the notations used throughout the paper, and then demonstrate how CAVs are generated as proposed by \cite{kim2018interpretability}. After that, we introduce the notations for traditional and post-hoc CBMs to familiarise the readers with these methodologies and their differences. Finally, the global and local concept importance notions are discussed. 

Let ${\cal X}$ be the set of images, ${\cal U}$ be the set of concept labels, and ${\cal Y} = \{1, 2, \cdots, K\}$ be the set of $K$ class labels. Let ${\cal S} = \{(\vect{X}_i, \vect {u{_i}}, \vect{y_i},{\Lambda}_i, {\cal P}_i) \ | \ \vect{X}_i \in {\cal X}, \vect{u}_i \in \mathcal{U}, \vect{y_i} \in {\cal Y}, i = 1, 2, \ldots, N\}$ be the training set with $N$ samples, where $\vect{u}_{i} \in \{0,1\}^L$ is the concept label vector with $L$ different concepts for image $\vect{X}_i \in \mathbb{R}^{M_1\times M_2\times M_3}$ ($M_3 = 3$ for RGB images),  $\vect{y_i} \in \mathbb{R}^K$ (a one-hot vector) denotes the class label of image $\vect{X}_i$, ${\Lambda}_i$ is the set containing the indexes of activated concepts for image $\vect{X}_i$ (i.e., the indexes of the components in $\vect{u}_{i}$ with value 1), and ${\cal P}_i = \{p_{i1}, \cdots, p_{iL}\}$ is the set holding centre pixel coordinates $p_{ij}$ of concept $j$ with $j = 1, \ldots, L$ for image $\vect{X}_i$. Let $f:\mathcal{X} \rightarrow \mathbb{R}^d$ be a $d$-dimensional feature extractor, which can be any trained DNN such as ResNet \cite{he2016deep} or VGG \cite{simonyan2014very}. From block \textcircled{a} in Figure~\ref{fig:pcbms}, we see that the feature vector $f(\vect{X}_i)$ consists of the post-GAP features (i.e., the features right after the GAP layer). Let $\vect{E}_i \in \mathbb{R}^{H \times W\times d}$ represent the pre-GAP feature maps (i.e., the features right before the GAP layer), where $H, W$ and $d$ denote the height, width and depth (i.e., the number of channels). The $k$-th channel of $\vect{E}_i$ is represented as $\vect{E}_i(:, :, k) \in \mathbb{R}^{H\times W}$.

\textbf{CAVs}. Following \citet{kim2018interpretability} and \citet{yuksekgonul2023posthoc}, for $j = 1, \ldots, L$, to generate the CAV $\vect{c}_j\in\mathbb{R}^{d}$ for the $j$-th concept, two sets of image embeddings through $f$ are needed, i.e., $\mathcal{N}^{\rm pos}_j$ for positive examples and $\mathcal{N}^{\rm neg}_j$ for negative ones. In detail, set $\mathcal{N}^{\rm pos}_j$ consists of embeddings of $N_p$ images (positive examples) that contain the $j$-th concept, and set $\mathcal{N}^{\rm neg}_j$ consists of embeddings of ${N}_n$
randomly chosen images (negative examples) that do not contain the concept. Sets $\mathcal{N}^{\rm pos}_j$ and $\mathcal{N}^{\rm neg}_j$ are then used to train an SVM with $\vect{c}_j$ being the obtained normal vector to the hyperplane separating sets $\mathcal{N}^{\rm pos}_j$ and $\mathcal{N}^{\rm neg}_j$. All together, these $L$ number of CAVs form a concept bank $\vect{C} = (\vect{c}_1, \cdots, \vect{c}_L)^\top \in \mathbb{R}^{L\times d}$. For an image $\vect{X}_i$, the feature vector $f(\vect{X}_i)$ is to be projected onto the concept space by $\vect{C}$, i.e., $\vect{C}f(\vect{X}_i)\in \mathbb{R}^L$, which is the concept value vector $\hat{\vect{u}}_i$ to be fed to the classifier.

\textbf{Traditional vs. post-hoc CBMs.} After the feature vector $f(\vect{X}_i)$ is obtained for image $\vect{X}_i$, the traditional CBMs predict concepts by the concept prediction block, while the post-hoc CBMs project the feature vector $f(\vect{X}_i)$ onto the concept space using the concept bank $\vect{C}$, see Figure \ref{fig:pcbms}. Let 
\begin{equation}
\hat{\vect{u}}_i = (\hat{u}_{i1}, \hat{u}_{i2}, \cdots, \hat{u}_{iL})^\top = g(f(\vect{X}_i))
\end{equation}
be the obtained concept vector for image $\vect{X}_i$ and $g$ be the projection function. Then, for traditional CBMs, as the ground-truth concept label vector ${\vect{u}_i}$ for image $\vect{X}_i$ is available, $g$ (i.e., the concept prediction block) is achieved/trained by minimising the binary cross-entropy loss function
$
\mathcal {L}_{g} = \sum_i  \mathcal {L}_{g}(\hat{\vect{u}}_i, \vect{u}_{i}).
$
In contrast, for post-hoc CBMs, where the ground-truth concept label vector $\vect{u}_i$ is not available, the obtained concept vector $\hat{\vect{u}}_i$ corresponding to $\vect{X}_i$ is directly obtained by setting 
$
g(f(\vect{X}_i)) = \vect{C}f(\vect{X}_i).
$

Finally, the obtained concept vector $\vect {\hat u_i}$ is used to predict the final classes via a single classification layer 
$
h:\mathbb{R}^{L} \rightarrow \cal{Y}
$
for both the traditional and post-hoc CBMs. In detail, 
$
h(\hat{\vect{u}}_i) = \vect{\theta}^\top\hat{\vect{u}}_i + b,
$
where $\vect{\theta} \in\mathbb{R}^{L\times K}$ holds the weights and $b$ is the bias. Function $h$ is trained for the final classification with the categorical cross-entropy loss function
$
\mathcal {L}_h = \sum_i  \mathcal {L}_h(\hat{\vect{y}}_i, \vect{y}_{i}),
$
where $\hat{\vect{y}}_i = h(g(f(\vect{X}_i)))$ is the class prediction of $\vect{X}_i$.

\textbf{Global vs. local concept importance.} After training the model is used to make a prediction for every test image, and then rank the concepts and present the highest $l$ of them as explanations. In this regard, it is crucial to differentiate between the global and local importance of concepts for a task as they may play key roles in different scenarios. 
Global importance is the overall effect of concepts for a given class. For instance, in the post-hoc CBMs setting, the classifier $h$ is a single layer with weights $\vect{\theta}$  mapping concept values to the final classes (also see the right of Figure \ref{fig:pcbms}) and each parameter of this layer is proposed as the \textit{global importance} of a concept that they weigh for an examined class. By analysing each parameter, say ${\theta}_{jk}$, one can assess the overall effect of the concept $j$ for class $k$. Moreover, tuning these parameters may allow the model to be debugged as proposed in \citet{yuksekgonul2023posthoc}.
The \textit{local importance} of concepts on the other hand is their influence on individual image predictions rather than on the entire class. The CBM and its variants focus on \textit{local concept interventions} \cite{kim2018interpretability}, which is the process of tweaking the predicted/projected concept values in $\hat{\vect{u}}_i$ at the concept bottleneck layer, i.e., \textcircled{d} in Figure \ref{fig:pcbms}, to flip a single class prediction when needed. An effective way to determine what concept values to intervene on is an active area of research \cite{pmlr-v235-steinmann24a, vandenhirtz2024stochastic, shin2023closer}. 

One, however, should note that the magnitude of the concept values at the bottleneck layer is not the same as the \textit{local importance}. This is because a class prediction score by $h$ is the $\vect{\theta}$-weighted sum of concept values, and the parameters of $\vect{\theta}$ may greatly increase or decrease the individual concept effects on the final classification. Therefore, defining the concept importance solely based on their values in $\hat{\vect{u}}_i$ is misleading. We will address this issue in our proposed methodology in Section \ref{sec:methodology}.

%-------------
\section{Proposed Methodology}
\label{sec:methodology}
%-------------
There is a significant gap in the field regarding the evaluation of the explainability power of the well-known concept-based methodologies. To fill this gap and assess the existence and correctness of the concepts given as highly important by XAI techniques, we propose our CoAM (concept activation mapping) framework (see \textcircled{f} in Figure \ref{fig:pcbms} for an overview), which allows concept visualisation. Moreover, we also propose the CGIM (concept global importance metric) to test the global concept alignment by XAI methods, the CEM (concept existence metric) to evaluate the existence of the concepts, and the CLM (concept location metric) to reveal whether the highly important concepts correspond to the correct regions in a given test image.

\subsection{Concept Activation Mapping}

We propose the CoAM framework, which generates concept activation maps revealing the parts of an image that correspond to the concepts. 
As we know, for post-hoc CBMs, the pre-GAP feature maps $\vect{E}_i \in \mathbb{R}^{H \times W \times d}$ (which contain spatial information) for the examined image $\vect{X}_i$ become the post-GAP feature vector $f(\vect{X}_i)$ after the GAP layer, which is then linked to the CAVs,  $\vect{c}_j = (c_{j1}, \cdots, c_{jd})^\top, j = 1, \ldots, L$. 

Our introduced concept activation maps, say $\vect{F}_{ij}$, for $\vect{X}_i$ corresponding to the $j$-th concept for $j = 1, \ldots, L$,  are calculated by
\begin{equation}
\label{CoAM}
\vect{F}_{ij} = \frac{1}{d}  \sum_{k=1}^d c_{jk} \vect{E}_i(:, :, k) \in \mathbb{R}^{H\times W}, 
\end{equation}
i.e., weighing the pre-GAP feature maps of $\vect{X}_i$ by the $j$-th CAV; see block \textcircled{c} in Figure \ref{fig:pcbms}. 
The CoAM framework is also summarised in Algorithm \ref{alg:CoAM}, with $\vect{F}_{i}$ being the output, where $\vect{F}_i(:, :, j) = \vect{F}_{ij}$ for $j = 1, 2, \ldots, L$.
Since the size of each $\vect{F}_{ij}$ is significantly smaller than that of $\vect{X}_i$, to visualise the concept activation maps in a better way and for localisation assessment, we upsample them to the original image size of $\vect{X}_i$, denoted by $\bar{\vect{F}}_{ij}$, and overlay them on $\vect{X}_i$. This will tell us what parts of the input image contribute to the individual concepts. Algorithm \ref{alg:heatmap} in the Appendix gives the details of the final feature visualisation pseudo-code.

\subsection{Concept Global Importance Metric}

We firstly introduce the \textit{global importance score} of concept $j$ for class $k$ as $\theta_{jk}$ [the $(j,k)$-th entry of $\vect{\theta}$], i.e., the weight in the classifier $h$ mapping the $j$-th concept to the $k$-th class, for $j = 1, 2, \ldots, L$ and $k = 1, 2, \ldots, K$. Let $\vect{V} \in\mathbb{R}^{L\times K}$ be the ground-truth concept matrix for all the classes provided by annotators, where the entry of its $j$-th row and $k$-th column ${V}_{jk}$ is the ground-truth value of the $j$-th concept for the $k$-th class. One might consider directly comparing ${\theta}_{jk}$ and  ${V}_{jk}$ for global evaluation of the correctness of ${\theta}_{jk}$. However, this is inappropriate because these values are on different scales; in particular, $\vect{V}$ contains values between $0$ and $1$, while $\vect{\theta}$ can take any real value as it represents layer weights. To address this issue, we propose to compare the entire $j$-th row vectors $\vect{\theta}(j,:)$ and $\vect{V}(j,:)$ by calculating their similarity for $j = 1, 2, \ldots, L$. 

Our {\it first type} CGIM is defined as 
\begin{equation} \label{CGIM-1}
\rho_j^{{\rm CGIM}_1} := \phi(\vect{\theta}(j,:), \vect{V}(j,:)), \ \ j = 1, 2, \ldots, L,
\end{equation}
where $\phi$ is the function for similarity calculation. In this paper, we use the cosine similarity (measuring the alignment between two vectors regardless of their magnitudes) for $\phi$. Therefore, $\rho_j^{{\rm CGIM}_1}$ is a similarity score between $-1$ and $1$ for the $j$-th concept. Ideally, $\rho_j^{{\rm CGIM}_1}$ is expected to be close to 1 if the obtained $\vect{\theta}(j,:)$ is meaningful. 

Analogous to the first type CGIM in Eqn \eqref{CGIM-1}, we also introduce the concept global explanations based on the average say $\hat{\vect{u}}^*_k$ of the concept vectors $\hat{\vect{u}}_i$ of $\forall \vect{X}_i \in {\cal X}_{\rm T}^k$, where ${\cal X}_{\rm T}^k$ is the set that consists of all the test images with correct predicted class $1\le k \le K$ and $|{\cal X}_{\rm T}^k| = N_{k}$. Then, form $\hat{\vect{U}}^* = (\hat{\vect{u}}^*_1, \hat{\vect{u}}^*_2, \cdots, \hat{\vect{u}}^*_K) \in \mathbb{R}^{L\times K}$, i.e., the obtained average concept matrix. Our {\it second type} CGIM is then defined as 
\begin{equation} \label{CGIM-2}
\rho_j^{{\rm CGIM}_2} := \phi(\hat{\vect{U}}^*(j,:), \vect{V}(j,:)), \ \ j = 1, 2, \ldots, L.
\end{equation}
If we consider both the weight matrix $\vect{\theta}$ and the obtained average concept matrix $\hat{\vect{U}}^*$, we have our {\it third type} CGIM, which is defined as 
\begin{equation} \label{CGIM-3}
\rho_j^{{\rm CGIM}_3} := \phi(\hat{\vect{U}}_{\vect{\theta}}^*(j,:), \vect{V}(j,:)), \ \ j = 1, 2, \ldots, L,
\end{equation}
where $\hat{\vect{U}}_{\vect{\theta}}^* = \vect{\theta}\odot\hat{\vect{U}}^*$ with $\odot$ being the pointwise multiplication operator.

The above proposed CGIM scores $\rho_j^{{\rm CGIM}_1}$, $\rho_j^{{\rm CGIM}_2}$, and $\rho_j^{{\rm CGIM}_3}$ are for each concept $1\le j \le L$. They can also be readily modified analogously so that we can calculate CGIM scores for each class $1\le k \le K$, i.e.,  
\begin{align} 
\rho_k^{{\rm CGIM}_1} & := \phi(\vect{\theta}(:,k), \vect{V}(:,k)), \\
\rho_k^{{\rm CGIM}_2} & := \phi(\hat{\vect{U}}^*(:,k), \vect{V}(:,k)), \\
\rho_k^{{\rm CGIM}_3} & := \phi(\hat{\vect{U}}_{\vect{\theta}}^*(:,k), \vect{V}(:,k)).
\end{align}

\subsection{Concept Existence Metric}

We now define the \textit{local importance score} of concept $j$ for class $k$ as ${\theta}_{jk} \hat{{u}}_{ij}$; note that $\hat{{u}}_{ij}$ is the obtained $j$-th  concept value of test image $\vect{X}_i$ and ${\theta}_{jk}$ is the weight in classifier $h$ linking the $j$-th  concept and the $k$-th class prediction. We rank the total $L$ concepts for test image $\vect{X}_i$ based on their contribution to the final classification prediction $k$ using the local importance score ${\theta}_{jk} \hat{{u}}_{ij}$, and let $\vect{q}_i = (q_{i1}, q_{i2}, \cdots, q_{iL})^\top$ represent the ranked indexes of the concepts for $\vect{X}_i$. Therefore, if $q_{is} = m$, it means the $m$-concept is ranked at the $s$ place for $s = 1, 2, \ldots, L$ based on the descending order of the magnitude of ${\theta}_{mk} \hat{{u}}_{im}$ among $\{{\theta}_{jk} \hat{{u}}_{ij}\}_{j=1}^L$.

Recall that ${\Lambda}_i$ is the set containing the indexes of activated concepts for image $\vect{X}_i$. Our CEM is defined as
\begin{equation}
\rho_l^{\rm CEM} := \frac{1}{l} \sum_{j=1}^l \vect{1}_{\Lambda_i}(q_{ij}),
\end{equation}
assessing if the first $l \le L$ concepts (i.e., the first $l$ components) in $\vect{q}_i$ exist in the examined image $\vect{X}_i$, where $\vect{1}_{\Lambda_i}$ is an indicator function defined as
\begin{equation}
\vect{1}_{\Lambda_i} (x) =
\begin{cases}
1, & {\rm if} \ x\in \Lambda_i; \\
0, & {\rm otherwise}.
\end{cases}
\end{equation}
Obviously, the CEM $\rho_l^{\rm CEM}$ is an accuracy score between 0 and 1 evaluating the existence of highly important concepts in image $\vect{X_i}$, thanks to the set ${\Lambda}_i$ containing the indexes of activated concepts. CEM reveals the reliability of explanations generated by a trained model; for example, $\rho_l^{\rm CEM} = 0$ means none of the $l$ highly important concepts exists in the examined image, whereas $\rho_l^{\rm CEM} = 1$ means all of the $l$ highly important concepts exist in the examined image.
We remark that $\rho_l^{\rm CEM}$ can also be obtained in the same manner by using ${\theta}_{jk}$ or $\hat{{u}}_{ij}$ instead of ${\theta}_{jk} \hat{{u}}_{ij}$ as the local importance score for comparison purpose.

\begin{algorithm}[!tb]
  \caption{Concept Activation Mapping (CoAM) }
  \label{alg:CoAM}
  \begin{algorithmic}[1]
    \STATE \textbf{Input:}  Pre-GAP feature maps $\vect{E}_i \in \mathbb{R}^{H \times W \times d}$ for $\vect{X}_i$, and concept bank $\vect{C}  \in \mathbb{R}^{L \times d}$
  %  \vspace{0.05in}
    
    \STATE \textbf{Output:}
    Concept activation map $\vect{F}_{i} \in \mathbb{R}^{H  \times W \times L}$  
    \vspace{0.05in}
%    \STATE \rule{\linewidth}{0.4pt} % Horizontal line
    %\State Initialize a zero matrix $\vect{F}_{i}$ for spatial projections 
    %\vspace{0.05in}
    
    \FOR{each concept $j$ in $\vect{C}$}
            %\State Reshape $\vect{c}_j$ to $d \times 1 \times 1$ for broadcasting
            \STATE Compute the weighted map $\vect{F}_{ij}$ with Eqn \eqref{CoAM}

            \STATE Set $\vect{F}_i(:, :, j) = \vect{F}_{ij}$
        
        \ENDFOR
    \STATE \textbf{return} $\vect{F}_i$

  \end{algorithmic}
\end{algorithm}

\subsection{Concept Location Metric}

After checking whether the obtained important concepts of image $\vect{X_i}$ exist in the ground-truth set ${\Lambda}_i$ with CEM and generating concept activation maps with CoAM, we now propose CLM to assess whether the obtained concepts of image $\vect{X_i}$ correspond to the correct region in $\vect{X_i}$. 

Note that this check could be rigorously done by calculating the intersection over union (IoU) score if a ground-truth segmentation map per concept is available. However, the absence of these ground-truth maps makes this way impractical. In contrast, it will be much easier to mark some pixels, e.g. the coordinate information of the centre pixel for each important semantic area in an image, and then link the coordinate information to each concept. One useful label available for this purpose is the coordinate information of the centre pixel for each concept, i.e., ${\cal P}_{i}$, for image $\vect{X}_i$. 
 
The proposed CLM checks whether the concept-wise activation heatmap $\bar{\vect{F}}_{ij}$ for concept $j$ generated by CoAM contains the ground-truth centre location $p_{ij}$. For the $l \le L$ most important concepts of $\vect{X}_i$ obtained in $\vect{q}_i$, our CLM is defined as
\begin{equation}
\rho_l^\text{CLM} := \frac{1}{l} \sum_{j=1}^l \vect{1}_{\Omega_{ij}}(p_{ij}),
\end{equation}
where $\Omega_{ij}$ is the visual region of concept $j$ of $\vect{X}_i$. Obviously, $\rho_l^\text{CLM}$ is an accuracy score between $0$ and $1$ evaluating the alignment between the obtained  individual concept heatmaps and their actual region in the image $\vect{X}_i$. In particular, $\rho_l^\text{CLM} = 0$ means none of the $l$ highly important concepts corresponds to the correct region in the image, whereas a $\rho_l^\text{CLM} = 1$ score means all the $l$ highly important concepts correspond to the correct region in the image. Finally, we remark that there are many ways to generate the visual region $\Omega_{ij}$. In this paper, we use thresholding on the concept-wise activation heatmap $\bar{\vect{F}}_{ij}$ with threshould $\tau$ to obtain $\Omega_{ij}$.

\section{Experiments}
\label{sec:experiments}

In this section, we present benchmark results and evaluate the performance of the post-hoc CBMs using our proposed metrics. The benchmark fine-grained bird classification dataset, \texttt{Caltech-UCSD Birds (CUB)} \cite{wah2011caltech}, with concept annotations such as \textit{wing color}, \textit{beak shape} and \textit{feather pattern} is employed for the experiments. It consists of 200 different classes and 112 binary concept labels for around $11,800$ images. Additionally, the central pixel locations of 12 different body parts are provided and used for concept localisation assessment by the proposed CLM. 
Following \citet{yuksekgonul2023posthoc}, we employ a ResNet-18 \cite{he2016deep} trained on the \texttt{CUB} dataset\footnote{The trained CUB model is available at \url{https://github.com/osmr/imgclsmob}.} as the feature extractor $f$. CAVs are calculated as explained in Section \ref{sec:methodology} to create a concept bank $\vect{C}$ (also see \textcircled{c} in Figure \ref{fig:pcbms}). Finally, a single layer $h$ with weights $\vect{\theta} \in \mathbb{R}^{112 \times 200}$ is trained for the classification.

\subsection{Post-hoc CBMs Reproduction}
By employing the same model as the feature extractor and following the same steps for CAVs and classifier training, we reproduce the results of post-hoc CBMs \cite{yuksekgonul2023posthoc} with various hyperparameter combinations. The details of the post-hoc CBMs reproduction is given in the Appendix.

\begin{figure}[!ht]
    \centering
    % First Row
    \subfloat{%
        \put(45,85){\small \underline{\textit{Class}}}
        \put(142,85){\small \underline{\textit{Concept}}}
        \includegraphics[width=0.45\linewidth]{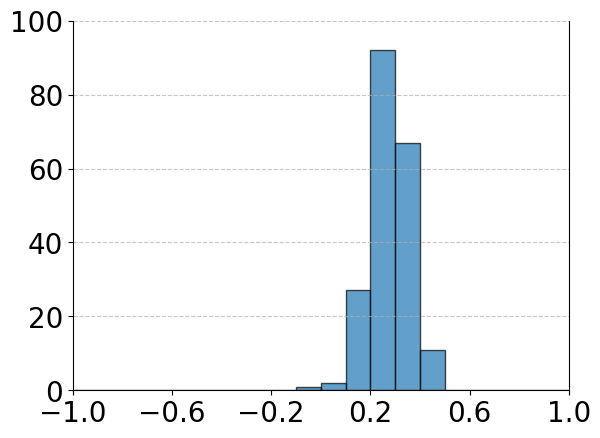}%
        \label{fig:class_hist}%
        \put(-65,-12){\small (a) $\rho_k^{{\rm CGIM}_1}$}
    }
    \subfloat{%
        \includegraphics[width=0.45\linewidth]{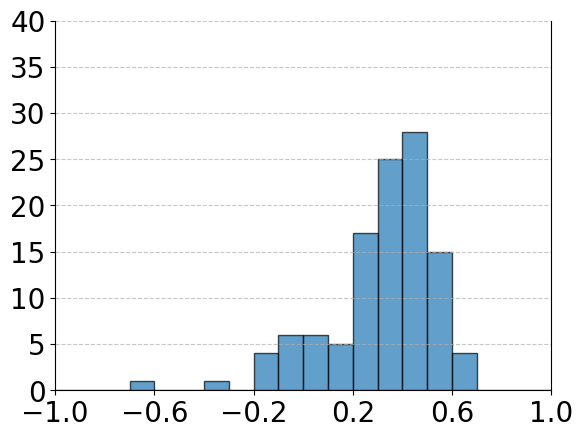}%
        \label{fig:concept_hist}%
         \put(-70,-12){\small (b) $\rho_j^{{\rm CGIM}_1}$}
    }
\vspace{-0.10in}
    
    % Second Row
    \subfloat{%
        \includegraphics[width=0.45\linewidth]{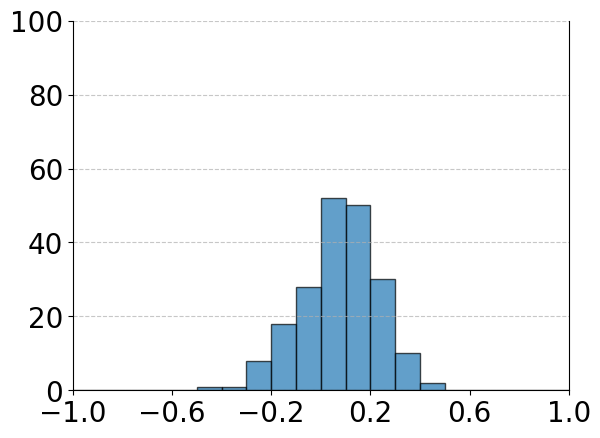}%
        \label{fig:class_avg}%
        \put(-65,-12){\small (c) $\rho_k^{{\rm CGIM}_2}$}
    }
    \subfloat{%
        \includegraphics[width=0.45\linewidth]{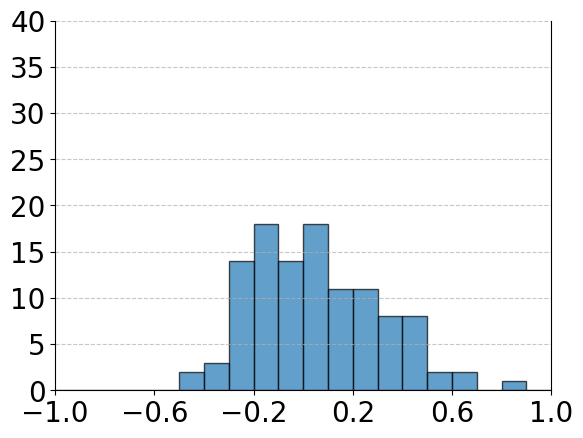}%
        \label{fig:concept_avg}%
        \put(-70,-12){\small (d) $\rho_j^{{\rm CGIM}_2}$}
    }

\vspace{-0.10in}

        % Third Row
    \subfloat{%
        \includegraphics[width=0.45\linewidth]{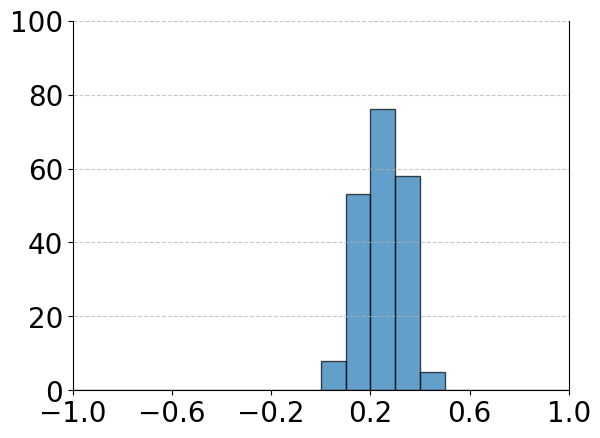}%
    \label{fig:class_weights_avg}%
    \put(-65,-12){\small (e) $\rho_k^{{\rm CGIM}_3}$}
    }
    \subfloat{%
        \includegraphics[width=0.45\linewidth]{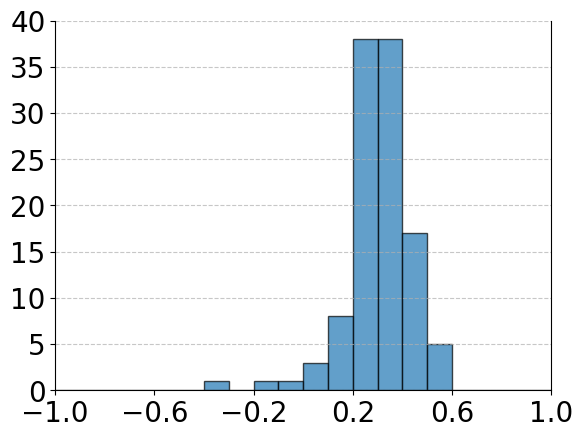}%
        \label{fig:concept_weights_avg}%
        \put(-65,-12){\small (f) $\rho_j^{{\rm CGIM}_3}$}
    }
    
    \caption{Histograms of the CGIM scores of the post-hoc CBMs. Plots on the left and right columns show the results for classes and concepts, respectively. A full list of the CGIM scores can be found in Table \ref{table:cos_sim_full} for the concepts and in Tables \ref{table:cos_sim_class} and \ref{table:cos_sim_class_continued} for the classes in the Appendix. }
    \label{fig:all_hist}
    \vspace{-0.05in}
\end{figure}

\subsection{Global Importance Evaluation}
\label{subsec:Global}

We now investigate the quality of the global explanations of the post-hoc CBMs. Recall that the entries of $\vect{\theta} \in \mathbb{R}^{112\times200}$ are considered as the global importance scores, determining the importance of a concept for an examined class. Ideally, these weights should closely align with human annotations in $\vect{V} \in \mathbb{R}^{112\times200}$, i.e., the so-called ground truth. 
Intuitively, we expect the CGIM scores $\rho_j^{{\rm CGIM}_1}$, $\rho_j^{{\rm CGIM}_2}$, and $\rho_j^{{\rm CGIM}_3}$ of $\vect{\theta}$, $\hat{\vect{U}}^*$, and $\hat{\vect{U}}_{\vect{\theta}}^*$ corresponding to $\vect{V}$ for each concept $1\le j \le 112$ (and analogously for each class $1\le k \le 200$) to be close to $1$ if the obtained $\vect{\theta}$, $\hat{\vect{U}}^*$, and $\hat{\vect{U}}_{\vect{\theta}}^*$ are meaningful.

The calculated CGIM scores of the post-hoc CBMs for each concept $1\le j \le 112$ and for each class $1\le k \le 200$ are respectively presented in Table \ref{table:cos_sim_full} and  Tables \ref{table:cos_sim_class} and \ref{table:cos_sim_class_continued} in the Appendix. To better visualise and interpret results, Figure~\ref{fig:all_hist} showcases the histograms of the obtained CGIM scores across a range between $-1$ (maximum dissimilarity) and $1$ (maximum similarity), regarding individual classes and concepts. Again, in an ideal scenario, it would be expected the CGIM scores $\rho_k^{{\rm CGIM}_1}$ and $\rho_j^{{\rm CGIM}_1}$ in the top row of Figure~\ref{fig:all_hist} to be a single prominent bar at the value of $1$, or at the very least, a clear accumulation of bars towards the right end of the histogram (approaching $1$), if the results of the post-hoc CBMs are meaningful/correct. Obviously, this is not the case. For example, the class and concept histograms in Figure \ref{fig:all_hist} (a)--(b) show that many bars are distributed across the range from $-1$ to $1$ with a noticeable number of values on the negative side, indicating a tendency towards negative correlation for some classes and concepts, which is contrary to the expected accumulation near $1$. The class and concept histograms in terms of $\rho_k^{{\rm CGIM}_2}$ and $\rho_j^{{\rm CGIM}_2}$ in Figure \ref{fig:all_hist} (c)--(d) and $\rho_j^{{\rm CGIM}_3}$ and $\rho_k^{{\rm CGIM}_3}$ in Figure \ref{fig:all_hist} (e)--(f) again disclose the same issue of the post-hoc CBMs.

A deeper analysis is also conducted through investigating the specific concepts and classes with significantly low or negative CGIM scores presented in Tables \ref{table:cos_sim_full}, \ref{table:cos_sim_class} and \ref{table:cos_sim_class_continued} in the Appendix.
For instance, the $\rho_j^{{\rm CGIM}_1}$ score for concept ($j = 51$) \textit{black eye colour} in Table \ref{table:cos_sim_full} is a large negative value, i.e., $-0.63$. Similarly, the $\rho_j^{{\rm CGIM}_1}$ score close to $0$ for class ($k=18$) \textit{spotted catbird} and class ($k=25$) \textit{pelagic cormorant} in Table \ref
{table:cos_sim_class} indicates that these classes share no similarities to their ground truth. For the first time, the low and negative valued CGIM scores occurring for many concepts and classes raise concerns about the reliability and quality of the explanations of the post-hoc CBMs.

\subsection{Concept Existence Evaluation}

After analysing the global importance evaluation based on classifier's weights and average concept predictions, we, in this section, focus on the local importance analysis. The first step in this regard is to assess the concept existence qualitatively and quantitatively.

\subsubsection{Qualitative observations} 
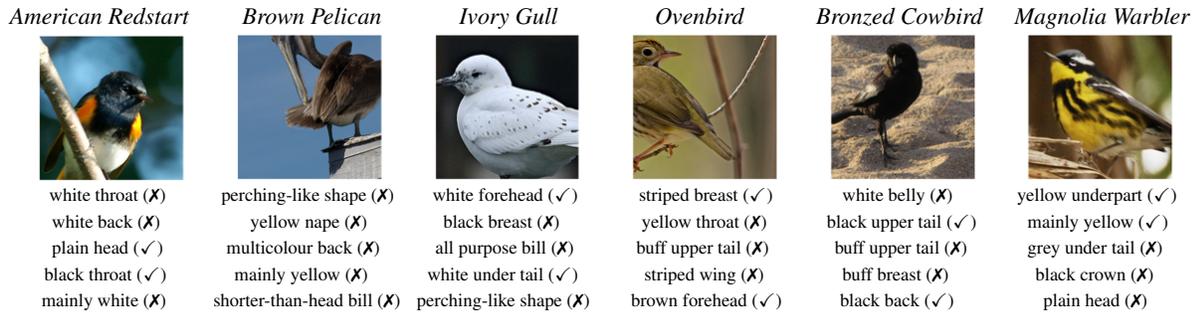
\begin{figure*}[htb]
\centering    \includegraphics[width=0.90\linewidth]{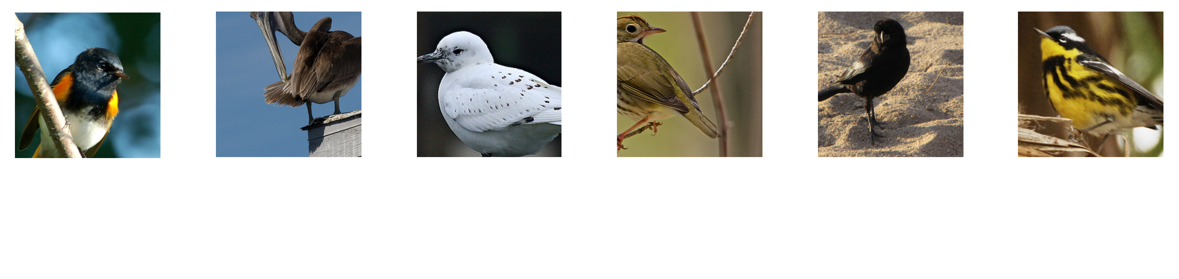}
    \put(-445, 104){\small \textit{American Redstart}}
    \put(-430, 37){\scriptsize white throat (\xmark)}
    \put(-429, 27){\scriptsize white back (\xmark)}
    \put(-429, 17){\scriptsize plain head (\checkmark)}
    \put(-432, 7){\scriptsize black throat (\checkmark)}
    \put(-433, -3){\scriptsize  mainly white (\xmark)}
    \put(-357, 104){\small \textit{Brown Pelican}}
    \put(-365, 37){\scriptsize perching-like shape (\xmark)}
    \put(-354, 27){\scriptsize yellow nape (\xmark)}
    \put(-363, 17){\scriptsize multicolour back (\xmark)}
    \put(-360, 7){\scriptsize mainly yellow (\xmark)}
    \put(-368, -3){\scriptsize shorter-than-head bill (\xmark)}
    \put(-275, 104){\small \textit{Ivory Gull}}
    \put(-285, 37){\scriptsize white forehead (\checkmark)}
    \put(-281, 27){\scriptsize black breast (\xmark)}
    \put(-284, 17){\scriptsize all purpose bill (\xmark)}
    \put(-287, 7){\scriptsize white under tail (\checkmark)}
    \put(-291, -3){\scriptsize perching-like shape (\xmark)}
    \put(-201, 104){\small \textit{Ovenbird}}
    \put(-207, 37){\scriptsize striped breast (\checkmark)}
    \put(-206, 27){\scriptsize yellow throat (\xmark)}
    \put(-208, 17){\scriptsize buff upper tail (\xmark)}
    \put(-205, 7){\scriptsize striped wing (\xmark)}
    \put(-210, -3){\scriptsize brown forehead (\checkmark)}
    \put(-140, 104){\small \textit{Bronzed Cowbird}}
    \put(-130, 37){\scriptsize white belly (\xmark)}
    \put(-136, 27){\scriptsize black upper tail (\checkmark)}
    \put(-133, 17){\scriptsize buff upper tail (\xmark)}
    \put(-130, 7){\scriptsize buff breast (\xmark)}
    \put(-131, -3){\scriptsize black back (\checkmark)}
    \put(-65, 104){\small \textit{Magnolia Warbler}}
    \put(-64, 37){\scriptsize yellow underpart (\checkmark)}
    \put(-60, 27){\scriptsize mainly yellow (\checkmark)}
    \put(-60, 17){\scriptsize grey under tail (\xmark)}
    \put(-57, 7){\scriptsize black crown (\xmark)}
    \put(-54, -3){\scriptsize plain head (\xmark)}
    \caption{Randomly selected test images from different classes and the top 5 most important concepts for their classification by the post-hoc CBMs. In particular, symbols \checkmark\ and \xmark\ are for concept existence and absence in the ground-truth label, respectively.}
    \label{fig:cem}
    \vspace{-0.05in}
\end{figure*}

When a set of concepts is presented as highly important for a prediction by a trained model, it is essential to qualitatively verify whether these concepts really exist in the image. In Figure \ref{fig:cem}, we present random images from the test set with the top $5$ most important concepts for their prediction outputted by the reproduced post-hoc CBMs. As shown in Figure \ref{fig:cem}, many of those highly important concepts do not actually exist in the given images. 
For instance, for an \textit{American Redstart} in the first column, the most important concept is given as \textit{white throat}; this is incorrect because the bird has a \textit{black throat}, which can be clearly seen in the input image. Similarly, for the \textit{brown pelican} image in the second column, the fifth most important concept is given as \textit{shorter than head bill}; this is not the case as the pelican has a much longer bill than its head.

\begin{table}[!tb]
\centering
\caption{Concept existence assessment of the reproduced post-hoc CBMs under CEM for the top $l$ most important concepts.}
\resizebox{0.40\textwidth}{!}
{\small 
\begin{tabular}{c||c||c|c|c} \hline Image & CEM based on & \textit{l = 1} & \textit{l = 3} & \textit{l = 5} \\ \hline \hline
                & $\theta_{jk}$       & {$39.2$} & {$37.9$}  &{$37.1$}  \\ 
Entire test set & $\hat{u}_{ij}$           & {$84.3$}  & {$80.1$}  &{$77.2$} \\
                &  $\theta_{jk} \hat{u}_{ij}$    & {$49.3$}  & {$44.3$}  &{$41.2$}  \\ \hline
                & $\theta_{jk}$      & {$48.5$}  & {$46.8$}  &{$44.8$}  \\ 
Correct class set & $\hat{u}_{ij}$   & $85.4$  & $82.1$  &$80.7$  \\
                & $\theta_{jk} \hat{u}_{ij}$     & $55.4$  & $49.1$  &$45.8$  \\ \hline

 \end{tabular}
}
\label{table:concept_acc}
\vspace{-0.05in}
\end{table}

\subsubsection{Quantitative test by CEM} We calculate the CEM score over the entire test set. The full results are presented in Table \ref{table:concept_acc} in terms of ranking the importance of the concepts based on i) the weights of the classifier ${\theta}_{jk}$, ii) the projected concept values $\hat{{u}}_{ij}$, and iii) their combination ${\theta}_{jk} \hat{{u}}_{ij}$, for the top $l$ most important concepts with $l$ set to $1$, $3$, and $5$. 
The results show that the CEM score based on $\hat{{u}}_{ij}$ is significantly higher than the others, which is intuitive at first glance as the highest values after concept projection are highly likely to be present in the ground-truth label. However, as detailed in Section \ref{sec:methodology}, the concept values in $\hat{\vect{u}}_i$ do not independently determine the final class prediction; instead, these values are weighted by their respective weights in $\vect{\theta}$, which can significantly alter their overall impact. Relying solely on the projected concept values in $\hat{\vect{u}}_i$ may therefore lead to misleading conclusions. Hence, we build our argument based on  ${\theta}_{jk} \hat{u}_{ij}$ rather than solely on $\hat{{u}}_{ij}$ or ${\theta}_{jk}$.  
Strikingly, as shown in Table \ref{table:concept_acc}, the single most important concept (i.e., when $l = 1$) only exists in the images around $55\%$ of the times when the image is correctly classified. This score drops to $49\%$ when the test is done on the entire test set. Moreover, the CEM score is even lower when $l$ is set to $3$ and $5$.

\subsection{Concept Localisation Evaluation}

% We now assess the concept localisation qualitatively using our proposed CoAM and quantitatively by our proposed CLM.

\subsubsection{Qualitative observations} By visualizing concept heatmaps for different concepts using our proposed CoAM, we identified several recurring patterns. Figure \ref{fig:examples} in the Appendix presents some examples of class and concept visualisation by using our CoAM. In many cases, the concept activation maps cover broad image regions, often extending beyond the expected concept areas. For instance, when detecting the concept \textit{grey leg} in a \textit{white breast nuthatch} image, the concept map covers the entire body of the bird rather than focusing on the specific region around the \textit{leg}, as shown in the first row in Figure \ref{fig:examples}. Moreover, many of the fine-grained concepts such as \textit{crown} or \textit{tail pattern} are often not correctly localised. For instance, the heatmap highlights a region around the \textit{leg} for \textit{blue crown} concept as given in the last row of Figure \ref{fig:examples}.

\subsubsection{Quantitative test by CLM} 
\label{sub:clm}

To be able to calculate the CLM score, the centre pixel coordinates for individual concepts are needed. In the \texttt{CUB} dataset, the centre pixel coordinates are only available for $12$ broader body parts such as \textit{beak, throat}, and \textit{leg}. Fortunately, most of the $112$ concepts are related to one of the $12$ body parts, allowing us to match each concept to its closest body part and hence exploit the corresponding body-part coordinates for concepts. For instance, we match the \textit{hooked seabird beak} concept with the \textit{beak} part and the \textit{solid wing} concept with the \textit{wing}; see Tables \ref{table:concept_groups} and \ref{tab:concept_groups_full} in the Appendix for details. We ignore concepts that are not related to a specific body part such as overall size, shape and colour information, which leaves us with $89$ out of $112$ concepts for the CLM evaluation.

Recall that after obtaining the activation map for the $j$-th concept $\bar{\vect{F}}_{ij}$ of $\vect{X}_i$, CLM checks if the centre pixel location $p_{ij}$ falls into the highest activated region $\Omega_{ij}$. Here $\Omega_{ij}$ is formed by the ${\alpha(M_1M_2)}/{12}$ number of pixels in terms of the largest pixel intensities in $\bar{\vect{F}}_{ij}$, where $\alpha$ is a hyperparameter that allows changing the region's size. For example, $\alpha = 1$ means the $1/12$ of the image is scanned, which is the size of a rough area for each of the 12 body parts such as \textit{beak, back} and \textit{throat} as given in Tables \ref{table:concept_groups} and \ref{tab:concept_groups_full}.

Table \ref{table:concept_local} gives the CLM scores for different choices of $\alpha$. For $\alpha = 1$, only $13.3\%$ of the time the centre pixel for a concept falls into the highly activated region $\Omega_{ij}$. Even increasing $\alpha$ to 6, which means half of the image is scanned, the centre pixels of the individual concepts are still not in the correct concept locations $41\%$ of the time.

\begin{table}[!tb]
\centering
\caption{Concept localisation assessment of the reproduced post-hoc CBMs under CLM for the top $l$ most important concepts.}
\label{table:concept_local}
\resizebox{0.42\textwidth}{!}
{\small 
\begin{tabular}{c||c||c|c|c} \hline Value $\alpha$ for $\Omega_{ij}$ & CLM based on & \textit{l = 1} & \textit{l = 3} & \textit{l = 5} \\ \hline \hline
                &  $\theta_{jk}$       & {$10.8$} & {$13.6$}  &{$13.8$}  \\ 
$\alpha = 1$ & $\hat{u}_{ij}$           & {$14.9$}  & {$14.6$}  &{$14.6$} \\
                & $\theta_{jk} \hat{u}_{ij}$   & {$13.3$}  & {$13.5$}  &{$12.9$}  \\ \hline
                & $\theta_{jk}$      & {$29.6$}  & {$30.1$}  &{$31.2$}  \\ 
$\alpha = 3$ & $\hat{u}_{ij}$   & $39.2$  & $33.7$  &$33.1$  \\
                & $\theta_{jk} \hat{u}_{ij}$     & $33.4$  & $32.2$  &$31.8$  \\ \hline
                & $\theta_{jk}$      & {$52.3$}  & {$50.8$}  &{$51.6$}  \\ 
$\alpha = 6$ & $\hat{u}_{ij}$   & $54.5$  & $56.4$  &$55.9$  \\
                & $\theta_{jk} \hat{u}_{ij}$     & $59.0$  & $55.0$  &$53.9$  \\ \hline

 \end{tabular}
}
\vspace{-0.05in}
\end{table}

\section{Discussion}
\label{sec:discussion}
%Leakage due to soft CBM as discussed in \cite{havasi2022addressing}..

Learning human-understandable concepts is a challenging task. Often concepts are highly
correlated with other features. For example, although hooves clearly relate to the feet of
animals, a group of hooved animals often share other common features (e.g., they are often
quadrupeds that feed on grass). Although these other features might help to identify the
concepts, it is not very helpful to be told that an important concept for determining that the
image represents a cow is hooves if the hooves are not visible in the image. It is therefore
important to check that the human-understandable concepts exist in the image and when they
exist the network is finding them in the correct location. By providing metrics and benchmarks we hope this will provide an important stimulus to develop models with
improved alignment.

Concept-based XAI methodologies showcase either global or local explainability of their proposed techniques, depending on their model setting. For instance, with the traditional CBMs, the quality of concept predictions can be assessed just like the final class predictions since the concept labels should be readily available for them to work in the first place. On the other hand, when concept labels are not available as in the post-hoc CBMs case, the main evaluation is on the classifier weights (i.e., $\vect{\theta})$ as the global explicator. This evaluation is further supported by model editing experiments. However, none of these experiments make a comparison between these weights and the ground-truth labels (i.e., $\vect{V}$ for the \texttt{CUB} dataset), hindering their reliability. Therefore, we propose CoAM and CGIM to visualise and evaluate the global explainability of concept-based XAI methodologies more rigorously and expose the alignment between the global concept explanations and ground-truth labels. Moreover, when the ground-truth concept labels are not available, for methods such as post-hoc CBMs, they are unable to do concept predictions and instead project concepts to the concept space, which would prevent a direct concept prediction evaluation even if the test set had ground-truth labels for each concept. Our CEM and CLM allows local concept evaluation regardless of the training settings of methodologies, i.e., whether they predict or project concepts to the concept space.

We should note, however, that the metrics we provide do not directly measure the usefulness of the
concepts as an explanation. Rather they act as sanity checks that the concepts are correctly
identified in the images. Also, success on the benchmark is not necessarily the top objective
of a network; for example, the motivation of the post-hoc CBMs \cite{yuksekgonul2023posthoc} was to provide a
low-cost means of building traditional CBMs. A traditional CBM that uses per-image annotation will
surely have a superior performance on our benchmark, but it may be too costly to train this
on other datasets.

Our findings raise important questions about the utility of current concept-based explanation methodologies in providing spatially grounded explanations for image-based tasks. While these models offer some degree of interpretability by linking decisions to human-understandable concepts, their failure to predict and localise concepts correctly can lead to misleading interpretations. This highlights the importance of more rigorous evaluation criteria such as CGIM, CEM and CLM and the development of models that prioritise both concept prediction accuracy and spatial interpretability.

\vspace{-0.08in}
\section{Conclusion}
\label{sec:conclusion}

In this paper, we proposed three novel metrics, i.e., CGIM, CEM and CLM, for concept-based XAI systems. CGIM provides a way to measure the global concept alignment ability of concept-based XAI techniques. CEM and CLM are introduced for local importance evaluation, testing if highly important concepts proposed by XAI techniques exist and can be correctly localised in a given test image, respectively. Employing these three metrics, we benchmarked post-hoc CBMs on the \texttt{CUB} dataset. Our experiments demonstrated significant limitations in current post-hoc methods, with many concepts and classes found to be weakly or even negatively correlated with their ground-truth labels by CGIM. Moreover, many concepts presented as highly important are not found to be present in test images by CEM, and their concept activations fail to align with the expected regions of the input images by CLM. As the field of XAI continues to evolve, it is essential to ensure that methods not only provide understandable concepts but also accurately predict and localise these concepts within input data. Future work may focus on improving both the concept prediction and spatial localisation capabilities of concept-based XAI methods, ensuring that they can offer reliable and interpretable insights across diverse applications.

\section{Impact Statement}

This paper presents work whose goal is to advance the field of Machine Learning. There are many potential societal consequences of our work, none which we feel must be specifically highlighted here.

%%%%%%%%%%%%%%%%%%%%%%%%%%%%%%%%%%%%%%%%%%%%%%%%%%%%%%%%%%%%%%%%%%%%%%%%%%%%%%%
%%%%%%%%%%%%%%%%%%%%%%%%%%%%%%%%%%%%%%%%%%%%%%%%%%%%%%%%%%%%%%%%%%%%%%%%%%%%%%%
% APPENDIX
%%%%%%%%%%%%%%%%%%%%%%%%%%%%%%%%%%%%%%%%%%%%%%%%%%%%%%%%%%%%%%%%%%%%%%%%%%%%%%%
%%%%%%%%%%%%%%%%%%%%%%%%%%%%%%%%%%%%%%%%%%%%%%%%%%%%%%%%%%%%%%%%%%%%%%%%%%%%%%%

\newpage
\appendix
\label{sec:appendix}
\onecolumn

%------------
\section{Related Work}
\label{sec:related}
%------------
This section recalls concept-based methodologies for XAI and examines existing variants of class activation mapping (CAM) highlighting the need for a dedicated approach to concept visualisation (which can be addressed by our CoAM).

{Network dissection \cite{bau2017network}} is one of the well-known concept-based approaches, where individual neurons in a network are examined to identify their correspondence to human-understandable concepts like \textit{edges, textures}, or \textit{objects}. By aligning neuron activations with segmentation-annotated images, 
%provided by for instance the \texttt{Broden} dataset, 
network dissection quantifies how well a model’s internal representations map to meaningful concepts. However, this method is computationally expensive and data-intensive, requiring large and richly labelled datasets to accurately associate neurons with interpretable concepts. Despite its valuable insights, these limitations have prompted the development of more efficient and flexible methods, such as CAVs and CBMs.

{Testing with CAVs (TCAV) framework \cite{kim2018interpretability}} introduced  CAVs to explain model predictions based on high-level human-interpretable concepts. CAVs represent directions in the latent space of a model corresponding to specific concepts, allowing for sensitivity analysis. By perturbing an input in the direction of a concept vector, TCAV measures how much the model's prediction depends on that specific concept, offering quantitative insights into the reliance on different concepts for a given task. TCAV has been applied in several fields to assess whether models depend on sensitive attributes like gender or race when making decisions. Recent adaptations have improved the computational efficiency and robustness of CAVs when applied to large-scale models \cite{ghorbani2019towards}. However, TCAV can only unveil the global effect of concepts on examined classes and not on individual samples. Therefore, it is unable to directly assess the concept predictions or provide spatial concept localisation for individual images.

CBMs \cite{koh2020concept} offers a significantly different approach to interpretability. They enforce that intermediate representations of the model correspond to human-understandable concepts, such as attributes (e.g., \textit{colour, shape, part}) of objects in an image. By constraining the model to predict based on these explicit concepts, CBMs inherently provide an interpretable mechanism for understanding decisions. This makes it easier to debug and correct errors by diagnosing the model’s performance on individual concepts. Recent work in CBMs has focused on improving robustness, especially when concept labels are noisy or incomplete. For instance, in Label-free CBMs \cite{oikarinen2023labelfree},  a method was proposed using unsupervised techniques to learn concept bottlenecks, thereby extending the applicability of CBMs to scenarios where manual labelling is expensive or impractical. Despite their interpretability, CBMs typically lack the ability to provide spatial visualisations, limiting their usefulness in tasks that require precise localisation of important concepts.

The multilevel XAI method in \cite{aysel2023multilevel} offers solutions for both expensive annotation needs and single-level output drawbacks of CBMs. The cost-effective solution to CBMs is achieved by only requiring class-wise concept annotations rather than per-image. Moreover, the multilevel XAI method provides concept-wise heatmaps by-product handling the single-level limitation of CBMs. To be more precise, different from other CBM approaches, the explanations by the multilevel XAI method are not only raw concept values, but also each concept comes with its saliency map that highlights the region in the image activated by that concept. The authors in \citet{aysel2023multilevel} have also shown the possibility of concept intervention on the input dimension, which is much more intuitive than the concept dimension. To give an example, in other CBMs, one may tweak the concept value, say, ``\textit{white}" at the bottleneck layer to flip the prediction, say, from polar bear to grizzly bear. In the multilevel XAI method, one can convert the white colour region in the image to brown to achieve the same flipping, which is more intuitive and reliable.

A breakthrough in visual explanations came with the introduction of CAM \cite{zhou2016learning}, which provides spatial localisation by computing class-specific activation maps that highlight the regions of an image most relevant for a given prediction. CAM operates by utilising the output of GAP layers in CNNs, enabling the generation of heatmaps that represent regions crucial for the final classification. This approach was generalised in Grad-CAM \cite{Selvaraju_2017_ICCV}, which makes use of the gradients flowing into the final convolutional layer to visualise where the model ``looks" when making a decision. Grad-CAM extends CAM to more general architectures without requiring specific layers like GAP. However, Grad-CAM does not always provide sharp localisation, especially when multiple objects are present in the image. Grad-CAM++ \cite{chattopadhay2018grad} addresses this limitation by refining the localisation to better handle multiple instances of objects, offering a more fine-grained interpretation. Further extensions include Score-CAM \cite{wang2020score}, which eliminates the dependency on gradients, instead using the activations themselves to weigh different regions of the input. This addresses some of the instability associated with gradient-based methods but comes with increased computational overhead. Other advancements like Ablation-CAM \cite{ramaswamy2020ablation} explore removing parts of the model and input to measure their impact on predictions, thus improving interpretability.

\section {Limitations}

There are drawbacks to the metrics CEM and CLM that we propose. The CEM can only be used on datasets where we have per-image annotations of the concepts for a test set (note that for our case, only a small number of concepts like the top $l \ll L$ are required per-image, and therefore is cheap). This 
limits its use to a very small number of datasets. Having a metric limited to one (or a small
number of datasets) runs the risk that models are developed that overfit to that particular
dataset. The CLM requires knowledge of the location of the concepts. In fact, the concept locations were not given and we had to do a “best guess”
approximation of whether the concepts found in the “saliency maps” overlap with the real
concept locations. It is also debatable whether the heatmaps we obtained by weighting the feature maps before doing GAP correctly capture the location of the concepts. In our judgment, this seems as fair an estimate of the position as
we can make. We feel there is considerable value in visualising the location of a concept
through the use of heatmaps. In \citet{aysel2023multilevel}, the authors built saliency maps for each
concept, but there they aligned each feature map to a concept which prevented
cross-contamination between concept locations. By providing visualisations of the parts of
the image that activated the concept, it made it much easier to assess the alignment of
concepts in that model. We have attempted to provide a similar visualisation for the
post-hoc CBMs \cite{yuksekgonul2023posthoc}, although as this is not part of the design of that model the visualisation
may not be perfect. Finally, reducing the assessment of alignment to a couple of numbers
loses a lot of fine-grain detail. As we illustrated, we can get a better understanding of the
failure of the network by examining the performance in more detail, for example, by plotting
histograms of the CBMs results to identify particularly poor concepts, or by visualising the
locations of the features to understand what concepts might be being learnt.

Despite those drawbacks, we believe that proposing a new benchmark for assessing concept
alignment has the potential to concentrate the effort of researchers on improving the
performance of concept-based XAI systems. As we have illustrated, the performance of
post-hoc CBMs is surprisingly poor. Without doing a systematic analysis of this alignment, it is
easy to overlook this problem and believe that an XAI system is more powerful than it
actually is. Our hope is that by introducing new metrics and  benchmarks we can improve the accuracy of
future concept-based XAI systems.

\section{Post-hoc CBMs Reproduction -- Details}

\begin{table}[!tb]
\centering
\caption{Classification accuracy of the reproduced post-hoc CBMs with different settings for the parameters $\lambda$, $N_p$, and $N_n$.}
\resizebox{0.37\textwidth}{!}
{\small 
\begin{tabular}{c||c|c} \hline \diagbox{$\lambda$}{$N_p=N_n$} & {\tt 50} & {\tt 100}  \\ \hline \hline
$0.001$ & $26.7$  & $52.2$        \\ 
$0.01$  & $34.1$  & $44.9$        \\ 
$0.1$   & $29.1$  & $41.5$        \\ 
$1$   & $25.5$  & \textbf{59.1} \\ 
$10$  & $25.3$  & $58.7$        \\   \hline
Traditional model w/o bottleneck & \multicolumn{2}{c}{$75.4$} \\ \hline 
\end{tabular}
}
\label{table:reproduced_results}
\end{table}

By employing the same model as the feature extractor and following the same steps for CAVs and classifier training, we reproduce the results of post-hoc CBMs \cite{yuksekgonul2023posthoc} with various hyperparameter combinations. 
There are two hyperparameters to tune during the SVM training for CAV learning, i.e., ${N}_p$ and ${N}_n$ (the number of positive and negative images per concept), which we set to $50$ and $100$, respectively. The other hyperparameter is the regularisation parameter $\lambda$ in SVM, which controls the trade-off between maximising the margin that separates classes and minimising classification errors on the training data. A low $\lambda$ value allows the model to prioritise a wider margin, even if some data points are misclassified, making the model more robust to noise and potentially improves its generalisation of new data. In contrast, a high $\lambda$ value forces the SVM to minimise the training error, making it less tolerant of misclassifications and resulting in a narrower margin. While a high $\lambda$ can lead to more accurate training performance, it may also increase the risk of overfitting, as the model becomes more sensitive to individual data points. Thus, $\lambda$ helps balance the SVM’s complexity and flexibility, impacting its ability to generalise well.

We train SVM with $\lambda$ values ranging from $0.001$ to $10$. Table \ref{table:reproduced_results} shows the classification accuracy of the classifier $h$ with various concept banks obtained by these hyperparameter combinations. For the experiments in the main paper, we employ the model with the best classification accuracy $59.1\%$, which is achieved when $N_p = N_n = 100$ and $\lambda = 1$. This result is very close to the accuracy $58.8\%$ reported in the seminal work \cite{yuksekgonul2023posthoc}. Note that there is more than $15\%$ accuracy loss in comparison to the traditional model, i.e., the one without a concept bottleneck (i.e., \textcircled{a} + \textcircled{e} in Figure \ref{fig:pcbms}), for the sake of obtaining an interpretable model via concept bottleneck.

\section {Concept Localisation Evaluation -- Figures and Tables}

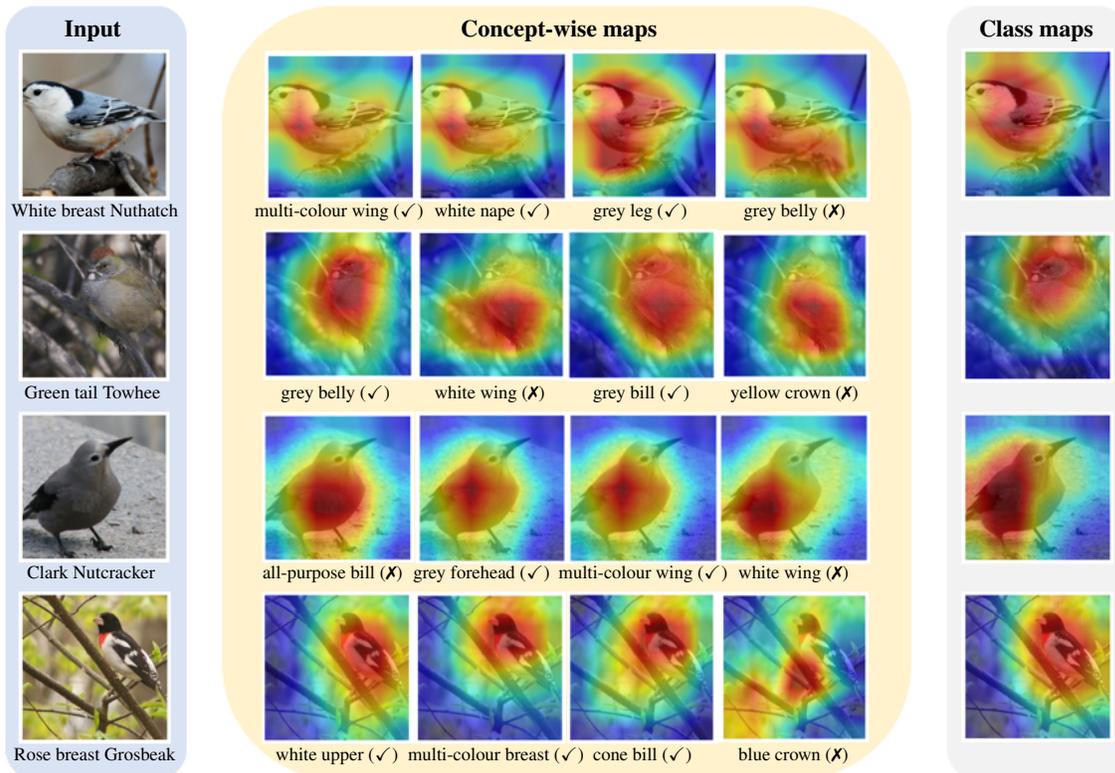
\begin{figure*}[htb]
    \centering
    \includegraphics[width=0.9\linewidth]{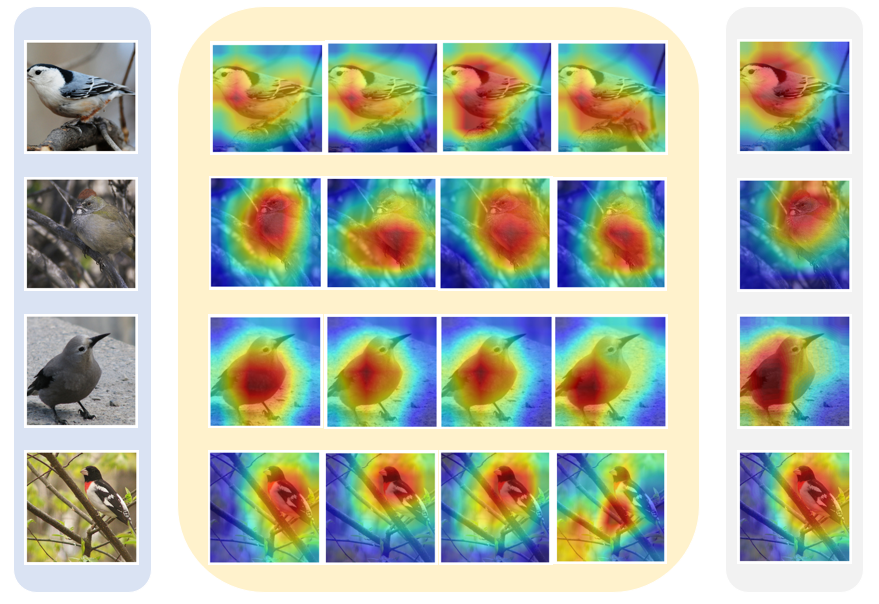}
    \put(-410, 287) {\small \textbf{Input}}
    \put(-260, 287) {\small \textbf{Concept-wise maps}}
    \put(-64, 287) {\small \textbf{Class maps}}
    \put(-430,219){\scriptsize White breast Nuthatch}
    \put(-338,219){\scriptsize multi-colour wing (\checkmark)}
    \put(-270,219){\scriptsize white nape (\checkmark)}
    \put(-210,219){\scriptsize grey leg (\checkmark)}
    \put(-153,219){\scriptsize grey belly (\xmark)}
    \put(-425,150){\scriptsize Green tail Towhee}
    \put(-328,150){\scriptsize grey belly (\checkmark)}
    \put(-270,150){\scriptsize white wing (\xmark)}
    \put(-210,150){\scriptsize grey bill (\checkmark)}
    \put(-158,150) {\scriptsize yellow crown (\xmark)}
    \put(-424, 82) {\scriptsize Clark Nutcracker}
    \put(-335,82){\scriptsize all-purpose bill (\xmark)}
    \put(-278,82){\scriptsize grey forehead (\checkmark)}
    \put(-223,82){\scriptsize multi-colour wing (\checkmark)}
    \put(-155,82){\scriptsize white wing (\xmark)}
    \put(-429, 13) {\scriptsize Rose breast Grosbeak}
    \put(-330,13){\scriptsize white upper (\checkmark)}
    \put(-280,13){\scriptsize multi-colour breast (\checkmark)}
    \put(-210,13){\scriptsize cone bill (\checkmark)}
    \put(-155,13){\scriptsize blue crown (\xmark)}
    \caption{Class and concept visualisation with our CoAM. All images (on the left) are correctly classified and their class-wise saliency maps are given on the right. The four most important concepts under CEM for the given classifications and their individual saliency maps are given in the middle. In particular, symbols \checkmark\ and \xmark\ are for concept existence and absence in the ground-truth label, respectively.}
    \label{fig:examples}
\end{figure*}

Figure \ref{fig:examples} presents some examples of class and concept visualisation by using our CoAM. The matching between the concept groups and the body parts for the \texttt{CUB} dataset is given in Tables \ref{table:concept_groups} and \ref{tab:concept_groups_full}.

\begin{table}[!htb]
\centering
\caption{The number of concepts, grouped based on their types, and mapped to the body parts (see Table \ref{tab:concept_groups_full} for more details).}
\label{table:concept_groups}
\resizebox{0.38\textwidth}{!}{
\begin{tabular}{|c|c|c|c|c|}
\hline
\diagbox{\textit{Part}}{\textit{Type}} & \textbf{Color} & \textbf{Pattern} & \textbf{Shape}  & \textbf{\textit{Total}} \\ \hline

\textbf{Back} & $6$  & $3$ & ---   & $9$ \\ \hline

\textbf{Beak} & $3$  & $2$ & $4$  & $9$ \\ \hline

\textbf{Belly} & $6$ & $1$ & ---  & $7$ \\ \hline

\textbf{Breast}& $6$ & $3$  &  --- & $9$ \\ \hline

\textbf{Crown} & $6$ & --- &  --- & $6$ \\ \hline

\textbf{Head}  & $6$ & $2$ & --- &$8$  \\ \hline

\textbf{Eye}  & $1$ & ---  & --- & $1$ \\ \hline

\textbf{Leg} & $3$  & --- & ---  & $3$ \\ \hline

\textbf{Wing} & $6$  & $4$ & $2$ & $12$   \\ \hline

\textbf{Nape} & $6$  & --- &  --- & $6$ \\ \hline

\textbf{Tail} & $10$ & $3$  & $1$ & $14$  \\ \hline

\textbf{Throat}& $5$  & --- &  --- & $5$ \\ \hline

\textbf{Others}& $18$  & --- &  $5$ & $23$ \\ \hline

\textbf{\textit{Total}}& $82$  & $18$  &  $12$ & $112$ \\ \hline

\end{tabular}
}
\end{table}

\begin{table*}[!htb]
\centering
\caption{Details of the concepts and the body parts they are mapped to.}
\label{tab:concept_groups_full}
\resizebox{!}{0.14\textwidth}{
\begin{tabular}{|c|c|c|c|}
\hline
\diagbox{\textit{Part}}{\textit{Type}} & \textbf{Color} & \textbf{Pattern (length for beak)} & \textbf{Shape}  \\ \hline

\textbf{Back}          & \textit{brown, grey, yellow, black, white, buff}  & \textit{solid, striped, multi-coloured} & ---    \\ \hline

\textbf{Beak}          & \textit{grey, black, buff}  & \textit{same-as-head, shorter-than-head} & \textit{dagger, hooked-seabird, all-purpose, cone}  \\ \hline

\textbf{Belly}         & \textit{brown, grey, yellow, black, white, buff}  & \textit{solid} & ---  \\ \hline

\textbf{Breast}        & \textit{brown, grey, yellow, black, white, buff} & \textit{solid, striped, multi-coloured}  &  ---  \\ \hline

\textbf{Crown}         & \textit{blue, brown, grey, yellow, black, white} & --- &  ---  \\ \hline

\textbf{Head}     & \textit{blue, brown, grey, yellow, black, white}  & \textit{eyebrow, plain} & ---   \\ \hline

\textbf{Eye}           & \textit{black} & ---  & ---    \\ \hline

\textbf{Leg}           & \textit{grey, black, buff}  & --- & ---   \\ \hline

\textbf{Wing}          & \textit{brown, grey, yellow, black, white, buff}  & \textit{solid, spotted, striped, multi-coloured} & \textit{rounded, pointed}   \\ \hline

\textbf{Nape}          & \textit{brown, grey, yellow, black, white, buff}  & --- &  ---  \\ \hline

\textbf{Tail}          & \textit{brown, grey, black, white, buff}  & \textit{solid, striped, multi-colored}  & \textit{notched}  \\ \hline

\textbf{Throat}        & \textit{grey, yellow, black, white, buff}  & --- &  ---  \\ \hline

\end{tabular}
}
\end{table*}

\section{Concept Visualisation Algorithm}

Concept activation maps in $\vect{F}_i$ for image $\vect{X}_i$ are obtained as summarised in Algorithm \ref{alg:CoAM}. These maps are at smaller resolutions since they are at a late layer of the trained DNN. Algorithm \ref{alg:heatmap} below demonstrates the steps to upsample these maps to the input size and we use them to mask the input image $\vect{X}_i$.

\begin{algorithm}[!ht]
  \caption{Feature Visualisation in CoAM}
  \label{alg:heatmap}
  \begin{algorithmic}[1]
    \STATE {Input:} 
    \begin{itemize} [noitemsep]
      \item Boolean flag \textit{coloured} for generating coloured heatmaps.
      \item Threshold value \textit{threshold} for binary heatmaps.
      \item Opacity level $\beta$ for superimposed heatmaps.
      \item Input image $\vect{X}_i \in \mathbb{R}^{M_1 \times M_2 \times M_3}$.
      \item Concept activation map $\vect{F}_i \in \mathbb{R}^{ H \times W \times L}$ of $\vect{X}_i$. \quad //  {$L$ is the number of concepts}
    \end{itemize}
   % \vspace{0.05in}
    
    \STATE {Output:} 
    Set of superimposed images $\bar {\mathcal{S}} \in \mathbb{R}^ {M_1 \times M_2 \times M_3 \times L}$.
    \vspace{0.00in}
    
    \noindent \rule{\linewidth}{0.4pt} % Horizontal line
    \STATE Initialize an empty list of superimposed images $\bar{\mathcal{S}}$
    \vspace{0.05in}

    %\For{each projection index $i$ in $\Lambda$}
    \FOR{each spatial projection map $j$ in $\vect{F}_i$}
        \STATE heatmap = \texttt{resize} ($\vect{F}_{ij}$, $(M_1, M_2))$ \quad // {Generate heatmap with size of $(M_1, M_2)$}
        % \vspace{0.05in}
        
        \IF{\textit{coloured}} %\COMMENT{Apply a colormap to the heatmap}
            \STATE jet\_heatmap = \texttt{apply\_colormap} (heatmap, ``jet") \quad // {Convert the heatmap to an RGB image}
            \STATE superimposed\_img = $\beta \cdot$ jet\_heatmap + $\vect{X}_i$ \quad // {Overlay heatmap on the original image $\vect{X}_i$}
            \STATE Append superimposed\_img to $\bar{\mathcal{S}}$
        % \vspace{0.05in}
        
        \ELSE %\COMMENT{Generate binary heatmap using the threshold value}
             
            \STATE binary\_heatmap = \texttt{binary\_threshold} (heatmap, \textit{threshold})
            \STATE superimposed\_img = $\vect{X}_i \odot$ binary\_heatmap \quad // {Overlay heatmap on the original image $\vect{X}_i$}
            \STATE Append superimposed\_img to $\bar {\mathcal{S}}$
            \vspace{0.05in}

        \ENDIF
    \ENDFOR
    
    \STATE {return} $\bar {\mathcal{S}}$

  \end{algorithmic}
\end{algorithm}

\section{CGIM Scores}

Figure \ref{fig:all_hist} presents the histograms of the CGIM scores. The full list of CGIM scores for every concept and every class is presented below in Table \ref{table:cos_sim_full} and in Tables \ref{table:cos_sim_class} and \ref{table:cos_sim_class_continued}, respectively.

\begin{table*}[!h]
\centering
\caption{Full list of CGIM scores for concepts in \texttt{CUB} dataset \cite{wah2011caltech} with reproduced post-hoc CBMs \cite{yuksekgonul2023posthoc}.}
\label{table:cos_sim_full}
\resizebox{!}{0.6\columnwidth}{
\begin{tabular}{|c|c|c|c|c|c|c|c|c|}
\hline
\diagbox{\textit{Concept}}{\textit{CGIM}} & $\rho_j^{{\rm CGIM}_1}$ & $\rho_j^{{\rm CGIM}_2}$ & $\rho_j^{{\rm CGIM}_3}$ &\diagbox{\textit{Concept}}{\textit{CGIM}} & $\rho_j^{{\rm CGIM}_1}$& $\rho_j^{{\rm CGIM}_2}$  & $\rho_j^{{\rm CGIM}_3}$    \\ \hline

1: Dagger beak & $0.54$   & $0.05$ & $0.41$ & 57: Yellow forehead colour  & $0.41$ &  $0.22$ & $0.47$  \\ \hline

2: Hooked seabird beak & $0.45$ & $0.08$ & $0.34$ & 58: Black forehead colour  & $0.39$ & $0.25$ & $0.35$\\ \hline

3: All-purpose beak & $0.37$  & $0.67$ & $0.46$ & 59: White forehead colour & $0.52$ & $-0.13$ & $0.07$   \\ \hline

4: Cone beak & $0.47$  & $0.23$ & $0.43$ & 60: Brown under tail colour & $-0.01$ & $-0.16$ & $0.23$ \\ \hline

5: Brown wing colour & $0.22$  & $-0.05$ & $0.28$ & 61: Grey under tail colour & $0.36$  & $-0.21$ & $0.30$ \\ \hline

6: Grey wing colour  & $-0.13$  & $-0.25$ & $0.31$ & 62: Black under tail colour & $-0.11$ & $0.35$ & $0.24$\\ \hline

7: Yellow wing colour  & $0.42$  & $-0.04$ & $0.42$ & 63: White under tail colour & $0.02$ & $-0.22$ & $0.23$ \\ \hline

8: Black wing colour & $0.42$   & $-0.02$ & $0.33$ & 64: Buff under tail colour & $0.46$  & $-0.21$ & $0.22$\\ \hline

9: White wing colour & $-0.05$ & $0.08$ & $0.30$ & 65: Brown nape colour & $0.22$  & $0.00$ & $0.39$ \\ \hline

10: Buff wing colour & $0.22$ & $0.12$ & $0.30$ & 66: Grey nape colour & $0.27$  & $0.34$ & $0.38$ \\ \hline

11: Brown upper-part colour & $0.09$   & $0.03$ & $0.26$  & 67: Yellow nape colour & $0.35$ & $0.23$ & $0.52$  \\ \hline

12: Grey upper-part colour & $-0.36$  & $0.16$ & $0.23$ & 68: Black nape colour & $0.37$  & $0.00$ & $0.20$ \\ \hline

13: Yellow upper-part colour & $0.46$  & $0.09$ & $ 0.45$ & 69: White nape colour & $0.38$ & $0.16$ & $0.28$ \\ \hline

14: Black upper-part colour & $-0.04$ & $0.48$ & $0.30$ & 70: Buff nape colour & $0.49$ & $-0.24$ & $0.14$ \\ \hline

15: White upper-part colour & $0.32$  & $-0.17$ & $0.23$ & 71: Brown belly colour & $0.56$ & $-0.15$ & $0.22$ \\ \hline

16: Buff upper-part colour & $0.27$   & $-0.13$ & $0.20$ & 72: Grey belly colour & $0.52$ & $-0.12$ & $0.41$ \\ \hline

17: Brown underpart colour & $0.47$  & $0.02$ & $0.36$ & 73: Yellow belly colour & $0.20$ & $0.43$ & $0.27$ \\ \hline

18: Grey underpart colour & $0.19$ & $0.04$ & $0.37$ & 74: Black belly colour & $0.48$ & $0.21$ & $0.46$ \\ \hline

19: Yellow underpart colour & $0.40$ & $0.55$ & $0.51$ & 75: White belly colour & $0.01$  & $0.33$ & $0.21$ \\ \hline

20: Black underpart colour & $0.55$ & $0.05$ & $0.29$ & 76: Buff belly colour & $0.48$ & $-0.31$ & $-0.07$ \\ \hline

21: White underpart colour & $-0.04$ & $0.03$ & $0.26$ & 77: Rounded wing shape & $0.05$  & $0.03$ & $0.38$ \\ \hline

22: Buff underpart colour & $0.18$   & $-0.10$ & $0.20$ & 78: Pointed wing shape & $0.51$  & $-0.16$ & $0.34$ \\ \hline

23: Solid breast pattern & $0.38$  & $0.39$ & $0.40$ & 79: Small size & $-0.09$ & $0.37$ & $0.28$ \\ \hline

24: Striped breast pattern & $0.33$  & $0.00$ & $0.33$ & 80: Medium size & $0.23$ & $0.06$ & $0.29$ \\ \hline

25: Multi-coloured breast pattern & $0.51$  & $-0.19$ & $0.25$ & 81: Very small size & $0.57$ & $-0.21$ & $0.26$ \\ \hline

26: Brown back colour & $0.43$  &$0.04$ & $0.23$ & 82: Duck-like shape & $0.42$ & $0.41$ & $0.57$ \\ \hline

27: Grey back colour & $0.38$  & $-0.21$ & $0.18$ & 83: Perching-like shape & $-0.18$ & $0.47$ & $0.11$ \\ \hline

28: Yellow back colour & $0.31$ & $0.03$ & $0.32$ & 84: Solid back pattern & $0.53$  & $-0.13$ & $0.20$ \\ \hline

29: Black back colour & $0.22$  & $0.60$ & $0.38$ &  85: Striped back pattern & $0.33$ & $-0.09$ & $0.33$ \\ \hline

30: White back colour & $0.34$  & $-0.23$ & $0.20$ & 86: Multi-coloured back pattern & $0.25$ & $-0.38$ & $0.39$ \\ \hline

31: Buff back colour & $0.01$   & $-0.24$ & $0.27$ & 87: Solid tail pattern & $0.64$ & $0.41$ & $0.45$ \\ \hline

32: Notched tail shape & $0.21$   & $-0.10$ & $0.31$ & 88: Striped tail pattern & $0.42$  & $-0.30$ & $0.21$ \\ \hline

33: Brown upper tail colour & $0.33$  & $-0.10$ & $0.23$ & 89: Multi-coloured tail pattern& $0.25$ &  $-0.41$ & $0.21$ \\ \hline

34: Grey upper tail colour & $0.08$   & $-0.19$ & $0.23$ & 90: Solid belly pattern & $0.35$  & $0.47$ & $0.37$ \\ \hline

35: Black upper tail colour & $0.44$  & $0.50$ & $0.50$ & 91: Brown primary colour & $0.12$ & $0.19$ & $0.26$ \\ \hline

36: White upper tail colour & $-0.13$ & $0.19$ & $0.25$ & 92: Grey primary colour & $0.45$ & $0.00$ & $0.27$ \\ \hline

37: Buff upper tail colour & $0.51$  & $-0.01$ & $0.39$ & 93: Yellow primary colour & $0.12$ & $0.44$ & $0.31$ \\ \hline

38: Head pattern eyebrow & $0.41$  & $-0.12$ & $0.37$ & 94: Black primary colour & $0.47$  & $0.11$ & $0.39$ \\ \hline

39: Head pattern plain & $0.67$  & $-0.05$ & $0.34$ & 95: White primary colour & $0.58$ & $-0.05$ & $0.26$ \\ \hline

40: Brown breast colour & $0.39$  & $0.19$ & $0.23$ & 96: Buff primary colour & $0.32$ & $-0.29$ & $0.26$ \\ \hline

41: Grey breast colour & $0.58$  & $-0.05$ & $0.35$ & 97: Grey leg colour & $0.55$  & $0.22$ & $0.43$ \\ \hline

42: Yellow breast colour & $0.47$ & $0.26$ & $0.46$ & 98: Black leg colour & $0.43$ & $-0.09$ & $0.29$ \\ \hline

43: Black breast colour & $0.22$  & $0.00$ & $0.25$ & 99: Buff leg colour & $0.17$  & $0.00$ & $0.27$ \\ \hline

44: White breast colour & $0.32$  & $-0.19$ & $0.30$ & 100: Grey bill colour & $0.43$ & $0.06$ & $0.41$ \\ \hline

45: Buff breast colour & $0.32$  & $-0.26$ & $0.17$ & 101: Black bill colour & $0.36$ & $0.39$ & $0.38$ \\ \hline

46: Grey throat colour & $0.38$  & $-0.11$ & $0.37$ & 102: Buff bill colour & $0.52$ &  $-0.43$ & $-0.13$ \\ \hline

47: Yellow throat colour & $0.26$ & $0.23$ & $0.29$ & 103: Blue crown colour & $0.37$ & $0.27$ & $0.50$ \\ \hline

48: Black throat colour & $0.52$ & $-0.03$ & $0.28$ & 104: Brown crown colour & $0.39$ & $0.16$ & $0.32$ \\ \hline

49: White throat colour & $0.45$ & $0.14$ & $0.32$ & 105: Grey crown colour & $0.39$ &  $-0.24$ & $0.19$ \\ \hline

50: Buff throat colour & $0.42$  & $-0.29$ & $0.22$ & 106: Yellow crown colour & $0.25$ & $0.20$ & $0.36$ \\ \hline

51: Black eye colour & $-0.63$ & $0.82$ & $-0.33$ & 107: Black crown colour & $0.45$  & $0.18$ & $0.39$ \\ \hline

52: Head size beak & $0.40$  & $0.26$ & $0.38$ & 108: White crown colour & $0.42$ & $-0.13$ & $0.01$ \\ \hline

53: Shorten than head size beak & $-0.08$ & $0.43$ & $0.23$ & 109: Solid wing pattern & $0.61$ & $0.39$ & $0.60$ \\ \hline

54: Blue forehead colour & $0.26$  & $0.36$ & $0.50$ & 110: Spotted wing pattern & $0.48$ & $0.04$ & $0.49$ \\ \hline

55: Brown forehead colour & $0.43$ & $-0.20$ & $0.32$ & 111: Striped wing pattern & $0.24$ & $-0.12$ & $0.37$ \\ \hline

56: Grey forehead colour & $0.62$ & $-0.20$ & $0.06$ & 112: Multi-coloured wing pattern & $0.27$ & $0.17$ & $0.40$ \\ \hline

\end{tabular}
}
\end{table*}

\begin{table*}
\centering
\caption{Full list of CGIM scores for classes in \texttt{CUB} dataset \cite{wah2011caltech} with reproduced post-hoc CBMs \cite{yuksekgonul2023posthoc}.}
\label{table:cos_sim_class}
\resizebox{!}{0.6\columnwidth}{
\begin{tabular}{|c|c|c|c|c|c|c|c|}
\hline
\diagbox{\textit{Class}}{\textit{CGIM}} & $\rho_k^{{\rm CGIM}_1}$ & $\rho_k^{{\rm CGIM}_2}$ & $\rho_k^{{\rm CGIM}_3}$ &\diagbox{\textit{Class}}{\textit{CGIM}} & $\rho_k^{{\rm CGIM}_1}$ & $\rho_k^{{\rm CGIM}_2}$ & $\rho_k^{{\rm CGIM}_3}$  \\ \hline

1: Black footed Albatross & $0.24$ & $-0.20$ & $0.03$ & 51: Horned Grebe  & $0.27$  & $-0.24$ & $0.24$ \\ \hline

2: Laysan Albatross &  $0.30$ & $0.04$ & $0.11$ & 52: Pied billed Grebe  & $0.32$ & $-0.12$ & $0.32$ \\ \hline

3: Sooty Albatross & $0.28$ & $-0.15$ & $0.17$ &  53: Western Grebe & $0.25$ & $-0.04$ & $0.24$\\ \hline

4: Groove billed Ani & $0.28$ & $0.37$ & $0.31$ & 54: Blue Grosbeak & $0.20$ & $0.20$  & $0.31$ \\ \hline

5: Crested Auklet & $0.25$ & $0.18$  & $0.26$ & 55: Evening Grosbeak & $0.40$ & $0.24$  & $0.35$ \\ \hline

6: Least Auklet & $0.26$ & $0.07$  & $0.14$ & 56: Pine Grosbeak & $0.14$ & $-0.01$   & $0.20$\\ \hline

7: Parakeet Auklet & $0.25$ & $0.12$  & $0.20$ & 57: Rose breasted Grosbeak & $0.31$ & $0.17$ & $0.31$  \\ \hline

8: Rhinoceros Auklet & $0.36$ & $0.00$ & $0.22$  & 58: Pigeon Guillemot & $0.36$ & $0.05$ & $0.20$ \\ \hline

9: Brewer Blackbird & $0.29$ & $0.21$ & $0.28$ & 59: California Gull & $0.27$ & $0.16$ & $0.24$ \\ \hline

10: Red winged Blackbird & $0.27$ & $0.36$ & $0.38$ & 60: Glaucous winged Gull & $0.37$ & $0.04$ & $0.26$ \\ \hline

11: Rusty Blackbird & $0.17$ & $-0.09$ & $0.32$  & 61: Heermann Gull & $0.26$ &$0.12$ & $0.32$  \\ \hline

12: Yellow headed Blackbird & $0.36$ & $0.32$ & $0.41$ & 62: Herring Gull & $0.27$ & $0.01$ & $0.17$ \\ \hline

13: Bobolink & $0.27$ & $0.23$ & $0.34$ & 63: Ivory Gull & $0.33$ & $0.33$ & $0.30$ \\ \hline

14: Indigo Bunting & $0.15$ & $0.16$ & $0.23$ & 64: Ring billed Gull & $0.34$ & $0.20$ & $0.29$ \\ \hline

15: Lazuli Bunting & $0.16$ & $-0.07$ & $0.27$ & 65: Slaty backed Gull & $0.25$ & $-0.04$ & $0.23$ \\ \hline

16: Painted Bunting & $0.28$ & $0.01$ & $0.36$ & 66: Western Gull & $0.16$ & $0.15$ & $0.20$ \\ \hline

17: Cardinal & $0.22$ & $0.00$ & $0.17$ & 67: Anna Hummingbird & $0.16$ & $-0.15$ & $0.17$ \\ \hline

18: Spotted Catbird & $-0.02$ &$-0.38$ & $0.14$ & 68: Ruby throated Hummingbird & $0.28$ &$-0.22$ & $0.17$ \\ \hline

19: Gray Catbird & $0.27$ & $0.17$ & $0.28$ & 69: Rufous Hummingbird & $0.19$ & $-0.09$ & $0.16$  \\ \hline

20: Yellow breasted Chat & $0.39$ & $0.01$ & $0.33$  & 70: Green Violetear & $0.20$ & $-0.14$ & $0.12$ \\ \hline

21: Eastern Towhee & $0.31$ & $0.00$ & $0.33$ 
 & 71: Long tailed Jaeger & $0.24$ & $-0.16$ & $0.13$ \\ \hline

22: Chuck will Widow & $0.23$ & $0.05$ & $0.32$ & 72: Pomarine Jaeger & $0.18$ & $-0.28$ & $0.36$ \\ \hline

23: Brandt Cormorant & $0.27$ & $0.02$ & $0.16$ & 73: Blue Jay & $0.30$ & $0.02$ & $0.36$ \\ \hline

24: Red faced Cormorant & $0.22$ & $0.15$ & $0.24$ & 74: Florida Jay & $0.32$ & $-0.06$ & $0.24$ \\ \hline

25: Pelagic Cormorant & $0.05$ & $0.06$ & $0.16$ & 75: Green Jay & $0.32$ & $0.13$ & $0.38$ \\ \hline

26: Bronzed Cowbird & $0.22$ & $0.19$ & $0.32$ & 76: Dark eyed Junco & $0.22$ & $-0.05$ & $0.20$ \\ \hline

27: Shiny Cowbird & $0.33$ & $0.25$ & $0.24$ & 77: Tropical Kingbird & $0.28$ & $0.13$ & $0.36$  \\ \hline

28: Brown Creeper & $0.30$ & $0.11$ & $0.27$ & 78: Gray Kingbird & $0.17$ & $0.06$ & $0.29$ \\ \hline

29: American Crow & $0.28$ & $0.38$ & $0.28$ & 79: Belted Kingfisher & $0.25$ & $0.01$   & $0.19$ \\ \hline

30: Fish Crow & $0.29$ & $0.44$  & $0.33$ & 80: Green Kingfisher & $0.27$ & $-0.04$  & $0.23$ \\ \hline

31: Black billed Cuckoo & $0.26$ & $0.04$  & $0.21$ & 81: Pied Kingfisher & $0.14$ & $0.41$  & $0.19$ \\ \hline

32: Mangrove Cuckoo & $0.23$ & $-0.22$  & $0.17$ & 82: Ringed Kingfisher & $0.21$ & $-0.18$  & $0.17$\\ \hline

33: Yellow billed Cuckoo & $0.26$  & $-0.02$  & $0.29$ & 83: White breasted Kingfisher & $0.39$ & $0.13$  & $0.28$ \\ \hline

34: Gray-crowned Rosy Finch & $0.33$  & $0.14$  & $0.21$ & 84: Red legged Kittiwake & $0.30$ & $0.17$  & $0.25$ \\ \hline

35: Purple Finch & $0.05$ & $0.00$  & $0.16$ & 85: Horned Lark & $0.31$ & $0.13$  & $0.23$ \\ \hline

36: Northern Flicker & $0.35$ & $0.02$  & $0.29$ & 86: Pacific Loon & $0.48$ & $-0.13$  & $0.17$ \\ \hline

37: Acadian Flycatcher & $0.32$ & $-0.15$  & $0.33$ & 87: Mallard & $0.27$ & $-0.14$  & $0.23$ \\ \hline

38: Great Crested Flycatcher & $0.37$ & $0.00$  & $0.30$ & 88: Western Meadowlark & $0.43$ &$0.05$  & $0.38$ \\ \hline

39: Least Flycatcher & $0.14$ & $-0.15$  & $0.16$ & 89: Hooded Merganser & $0.46$ & $-0.03$  & $0.38$ \\ \hline

40: Olive sided Flycatcher & $0.21$ & $-0.20$  & $0.25$ & 90: Red breasted Merganser & $0.24$ & $-0.10$  & $0.29$ \\ \hline

41: Scissor tailed Flycatcher & $0.25$ & $-0.01$  & $0.25$ & 91: Mockingbird & $0.22$ & $-0.03$  & $0.10$ \\ \hline

42: Vermilion Flycatcher & $0.12$ & $-0.01$  & $0.07$ & 92: Nighthawk & $0.32$ & $-0.12$  & $0.26$ \\ \hline

43: Yellow bellied Flycatcher & $0.37$ & $-0.22$  & $0.18$ & 93: Clark Nutcracker & $0.32$ & $0.20$  & $0.31$ \\ \hline

44: Frigatebird & $0.25$ & $0.04$  & $0.18$ & 94: White breasted Nuthatch & $0.24$ & $0.11$  & $0.34$ \\ \hline

45: Northern Fulmar & $0.14$ & $0.12$  & $0.16$ & 95: Baltimore Oriole & $0.27$ & $0.07$  & $0.30$ \\ \hline

46: Gadwall & $0.32$ & $-0.04$  & $0.36$ & 96: Hooded Oriole & $0.27$ & $0.17$ & $0.19$ \\ \hline

47: American Goldfinch & $0.38$ & $0.37$ & $0.37$ & 97: Orchard Oriole & $0.30$ & $0.08$ & $0.17$ \\ \hline

48: European Goldfinch & $0.45$ & $0.13$ & $0.32$ & 98: Scott Oriole & $0.43$  & $0.23$ & $0.43$ \\ \hline

49: Boat tailed Grackle & $0.32$ & $0.08$ & $0.25$ & 99: Ovenbird & $0.23$ & $0.00$ & $0.13$ \\ \hline

50: Eared Grebe & $0.26$ & $-0.10$ & $0.14$ &  100: Brown Pelican & $0.22$ & $-0.45$ & $0.06$ \\ \hline

\end{tabular}
}
\end{table*}

\begin{table*}
\centering
\caption{Full list of CGIM scores for classes in \texttt{CUB} dataset \cite{wah2011caltech} with reproduced post-hoc CBMs \cite{yuksekgonul2023posthoc} (Table \ref{table:cos_sim_class} continued).}
\label{table:cos_sim_class_continued}
\resizebox{!}{0.55\columnwidth}{
\begin{tabular}{|c|c|c|c|c|c|c|c|}
\hline
\diagbox{\textit{Class}}{\textit{CGIM}} & $\rho_k^{{\rm CGIM}_1}$ & $\rho_k^{{\rm CGIM}_2}$ & $\rho_k^{{\rm CGIM}_3}$ &\diagbox{\textit{Class}}{\textit{CGIM}} & $\rho_k^{{\rm CGIM}_1}$ & $\rho_k^{{\rm CGIM}_2}$ & $\rho_k^{{\rm CGIM}_3}$ \\ \hline

101: White Pelican & $0.19$ & $0.03$ & $0.05$ &  151: Black-capped Vireo & $0.25$  & $0.02$ & $0.27$ \\ \hline

102: Western Wood Pewee & $0.26$ & $0.05$ & $0.23$ & 152: Blue-headed Vireo & $0.35$  & $0.03$ & $0.30$ \\ \hline

103: Sayornis & $0.25$ & $-0.16$ & $0.13$ & 153: Philadelphia Vireo & $0.19$ & $0.00$ & $0.20$ \\ \hline

104: American Pipit & $0.27$ & $0.09$ & $0.27$ & 154: Red-eyed Vireo & $0.20$  & $0.00$& $0.32$ \\ \hline

105: Whip poor will & $0.24$ & $-0.21$ & $0.34$ & 155: Warbling Vireo & $0.34$ & $0.10$ & $0.14$ \\ \hline

106: Horned Puffin & $0.37$ & $0.24$ & $0.35$ & 156: White-eyed Vireo & $0.31$ & $0.05$ & $0.25$ \\ \hline

107: Common Raven & $0.32$ & $0.24$ & $0.33$ & 157: Yellow-throated Vireo & $0.21$ & $0.16$ & $0.13$ \\ \hline

108: White-necked Raven & $0.35$ & $0.22$ & $0.35$ & 158: Bay-breasted Warbler & $0.32$ & $-0.12$ & $0.22$ \\ \hline

109: American Redstart & $0.28$ & $0.12$ & $0.24$ & 159: Black-and-white Warbler & $0.35$ & $0.15$ & $0.30$ \\ \hline

110: Geococcyx & $0.24$ & $-0.10$ & $0.24$ & 160: Black-throated Blue Warbler & $0.26$  & $0.02$ & $0.27$ \\ \hline

111: Loggerhead Shrike & $0.33$ & $0.28$ & $0.31$ & 161: Blue-winged Warbler & $0.26$  & $0.36$ & $0.35$ \\ \hline

112: Great Grey Shrike & $0.28$ & $0.19$ & $0.31$ & 162: Canada Warbler & $0.27$ & $0.06$ & $0.17$ \\ \hline

113: Baird's Sparrow & $0.24$ &$0.12$ & $0.31$ & 163: Cape May Warbler & $0.28$ & $-0.05$ & $0.30$ \\ \hline

114: Black-throated Sparrow & $0.33$ &$0.08$ & $0.22$ & 164: Cerulean Warbler & $0.10$ & $-0.10$ & $0.21$ \\ \hline

115: Brewer's Sparrow & $0.25$ &$0.23$ & $0.12$ & 165: Chestnut-sided Warbler & $0.27$  & $0.02$ & $0.35$ \\ \hline

116: Chipping Sparrow & $0.28$ &$0.02$ & $0.21$ & 166: Golden-winged Warbler & $0.40$  & $0.20$ & $0.39$ \\ \hline

117: Clay-colored Sparrow & $0.22$ & $0.07$ & $0.16$ & 167: Hooded Warbler & $0.26$  & $0.17$  & $0.43$ \\ \hline

118: House Sparrow & $0.36$ & $0.11$ & $0.24$ & 168: Kentucky Warbler & $0.12$ & $0.17$ & $0.14$ \\ \hline

119: Field Sparrow & $0.14$ & $0.07$ & $0.19$ & 169: Magnolia Warbler & $0.41$ & $0.19$ & $0.35$ \\ \hline

120: Fox Sparrow & $0.22$ & $0.10$ & $0.22$ & 170: Mourning Warbler & $0.33$ & $0.06$ & $0.24$ \\ \hline

121: Grasshopper Sparrow & $0.30$ & $0.08$ & $0.34$ & 171: Myrtle Warbler & $0.34$ & $0.01$ & $0.22$ \\ \hline

122: Harris's Sparrow & $0.22$ & $-0.01$ & $0.31$ & 172: Nashville Warbler & $0.27$ & $0.18$ & $0.33$ \\ \hline

123: Henslow's Sparrow & $0.34$ & $0.10$ & $0.27$ & 173: Orange-crowned Warbler & $0.21$ & $0.00$ & $0.18$ \\ \hline

124: Le Conte's Sparrow & $0.36$ & $0.09$ & $0.29$ & 174: Palm Warbler & $0.18$ & $-0.05$ & $0.09$ \\ \hline

125: Lincoln Sparrow & $0.28$ & $0.22$ & $0.21$ & 175: Pine Warbler & $0.23$ & $0.23$ & $0.19$ \\ \hline

126: Nelson's Sharp-tailed Sparrow & $0.31$ & $-0.13$ & $0.17$ & 176: Prairie Warbler & $0.20$ & $0.16$ & $0.24$ \\ \hline

127: Savannah Sparrow & $0.35$ & $0.13$ & $0.32$ & 177: Prothonotary Warbler & $0.33$ & $0.37$ & $0.42$ \\ \hline

128: Seaside Sparrow & $0.15$ & $-0.24$ & $0.07$ & 178: Swainson's Warbler & $0.31$ & $0.05$ & $0.20$ \\ \hline

129: Song Sparrow & $0.37$ & $0.20$ & $0.29$ & 179: Tennessee Warbler & $0.17$ & $0.00$ & $0.20$ \\ \hline

130: Tree Sparrow & $0.38$ & $0.10$ & $0.20$ & 180: Wilson's Warbler & $0.20$ & $0.25$ & $0.34$ \\ \hline

131: Vesper Sparrow & $0.16$ & $0.12$ & $0.17$ & 181: Worm-eating Warbler & $0.33$ & $0.02$ & $0.32$ \\ \hline

132: White-crowned Sparrow & $0.40$ & $0.12$ & $0.33$ & 182: Yellow Warbler & $0.33$ & $0.37$ & $0.34$ \\ \hline

133: White-throated Sparrow & $0.18$ & $0.00$ & $0.27$ & 183: Northern Waterthrush & $0.23$ & $0.08$ & $0.24$ \\ \hline

134: Cape Glossy Starling & $0.40$ &$0.14$ & $0.34$ & 184: Louisiana Waterthrush & $0.27$ & $0.01$ & $0.19$ \\ \hline

135: Bank Swallow & $0.17$ &$-0.15$ & $0.19$ & 185: Bohemian Waxwing & $0.29$  & $0.23$ & $0.28$ \\ \hline

136: Barn Swallow & $0.37$ &$0.09$ & $0.22$ & 186: Cedar Waxwing & $0.40$ & $0.12$ & $0.25$ \\ \hline

137: Cliff Swallow & $0.29$ &$-0.13$ & $0.12$ & 187: American Three-toed Woodpecker & $0.32$ & $0.14$ & $0.20$ \\ \hline

138: Tree Swallow & $0.33$ &$0.17$ & $0.30$ & 188: Pileated Woodpecker & $0.31$ & $0.27$ & $0.26$ \\ \hline

139: Scarlet Tanager & $0.30$ &$0.10$ & $0.09$ & 189: Red-bellied Woodpecker & $0.20$ & $0.10$ & $0.27$ \\ \hline

140: Summer Tanager & $0.21$ &$0.10$ & $0.16$ & 190: Red-cockaded Woodpecker & $0.28$  & $0.26$ & $0.26$ \\ \hline

141: Arctic Tern & $0.23$  &$0.17$ & $0.26$ & 191: Red-headed Woodpecker & $0.39$ & $0.22$ & $0.33$ \\ \hline

142: Black Tern & $0.22$ &$-0.07$ & $0.27$ & 192: Downy Woodpecker & $0.27$ & $0.24$ & $0.27$ \\ \hline

143: Caspian Tern & $0.12$ &$0.20$ & $0.26$ & 193: Bewick Wren & $0.20$ & $0.25$ & $0.30$ \\ \hline

144: Common Tern & $0.17$ &$0.11$ & $0.26$ & 194: Cactus Wren & $0.39$ & $0.17$ & $0.29$ \\ \hline

145: Elegant Tern & $0.22$ & $0.23$& $0.25$ & 195: Carolina Wren & $0.33$ & $0.27$ & $0.30$ \\ \hline

146: Forsters Tern & $0.17$ & $0.25$ & $0.24$ & 196: House Wren & $0.26$ & $0.30$ & $0.28$ \\ \hline

147: Least Tern & $0.38$ & $0.20$ & $0.33$ & 197: Marsh Wren & $0.33$ & $0.07$ & $0.21$ \\ \hline

148: Green tailed Towhee & $0.30$ & $0.06$ & $0.27$ & 198: Rock Wren & $0.24$ & $0.19$ & $0.22$ \\ \hline

149: Brown Thrasher & $0.34$ &$0.10$ & $0.30$ & 199: Winter Wren & $0.20$ & $0.32$ & $0.16$ \\ \hline

150: Sage Thrasher & $0.37$ &$0.12$ & $0.25$ & 200: Common Yellowthroat & $0.30$ & $-0.04$ & $0.30$ \\ \hline 

\end{tabular}
}
\end{table*}

%%%%%%%%%%%%%%%%%%%%%%%%%%%%%%%%%%%%%%%%%%%%%%%%%%%%%%%%%%%%%%%%%%%%%%%%%%%%%%%
%%%%%%%%%%%%%%%%%%%%%%%%%%%%%%%%%%%%%%%%%%%%%%%%%%%%%%%%%%%%%%%%%%%%%%%%%%%%%%%


\begin{thebibliography}{24}
\providecommand{\natexlab}[1]{#1}
\providecommand{\url}[1]{\texttt{#1}}
\expandafter\ifx\csname urlstyle\endcsname\relax
  \providecommand{\doi}[1]{doi: #1}\else
  \providecommand{\doi}{doi: \begingroup \urlstyle{rm}\Url}\fi

\bibitem[Ali et~al.(2023)Ali, Abuhmed, El-Sappagh, Muhammad, Alonso-Moral, Confalonieri, Guidotti, Del~Ser, D{\'\i}az-Rodr{\'\i}guez, and Herrera]{ali2023explainable}
Ali, S., Abuhmed, T., El-Sappagh, S., Muhammad, K., Alonso-Moral, J.~M., Confalonieri, R., Guidotti, R., Del~Ser, J., D{\'\i}az-Rodr{\'\i}guez, N., and Herrera, F.
\newblock Explainable artificial intelligence (xai): What we know and what is left to attain trustworthy artificial intelligence.
\newblock \emph{Information Fusion}, 99:\penalty0 101805, 2023.

\bibitem[Aysel et~al.(2023)Aysel, Cai, and Prugel-Bennett]{aysel2023multilevel}
Aysel, H.~I., Cai, X., and Prugel-Bennett, A.
\newblock {Multilevel Explainable Artificial Intelligence: Visual and Linguistic Bonded Explanations}.
\newblock \emph{IEEE Transactions on Artificial Intelligence}, 2023.

\bibitem[Bau et~al.(2017)Bau, Zhou, Khosla, Oliva, and Torralba]{bau2017network}
Bau, D., Zhou, B., Khosla, A., Oliva, A., and Torralba, A.
\newblock Network dissection: Quantifying interpretability of deep visual representations.
\newblock In \emph{Proceedings of the IEEE Conference on Computer Vision and Pattern Recognition}, pp.\  6541--6549, 2017.

\bibitem[Buhrmester et~al.(2021)Buhrmester, M{\"u}nch, and Arens]{buhrmester2021analysis}
Buhrmester, V., M{\"u}nch, D., and Arens, M.
\newblock Analysis of explainers of black box deep neural networks for computer vision: A survey.
\newblock \emph{Machine Learning and Knowledge Extraction}, 3\penalty0 (4):\penalty0 966--989, 2021.

\bibitem[Chattopadhay et~al.(2018)Chattopadhay, Sarkar, Howlader, and Balasubramanian]{chattopadhay2018grad}
Chattopadhay, A., Sarkar, A., Howlader, P., and Balasubramanian, V.~N.
\newblock {Grad-cam++: Generalized gradient-based visual explanations for deep convolutional networks}.
\newblock In \emph{2018 IEEE Winter Conference on Applications of Computer Vision (WACV)}, pp.\  839--847. IEEE, 2018.

\bibitem[Ghorbani et~al.(2019)Ghorbani, Wexler, Zou, and Kim]{ghorbani2019towards}
Ghorbani, A., Wexler, J., Zou, J.~Y., and Kim, B.
\newblock Towards automatic concept-based explanations.
\newblock \emph{Advances in Neural Information Processing Systems}, 32, 2019.

\bibitem[Goebel et~al.(2018)Goebel, Chander, Holzinger, Lecue, Akata, Stumpf, Kieseberg, and Holzinger]{goebel2018explainable}
Goebel, R., Chander, A., Holzinger, K., Lecue, F., Akata, Z., Stumpf, S., Kieseberg, P., and Holzinger, A.
\newblock {Explainable AI: the new 42?}
\newblock In \emph{International Cross-domain Conference for Machine Learning and Knowledge Extraction}, pp.\  295--303. Springer, 2018.

\bibitem[Hassija et~al.(2024)Hassija, Chamola, Mahapatra, Singal, Goel, Huang, Scardapane, Spinelli, Mahmud, and Hussain]{hassija2024interpreting}
Hassija, V., Chamola, V., Mahapatra, A., Singal, A., Goel, D., Huang, K., Scardapane, S., Spinelli, I., Mahmud, M., and Hussain, A.
\newblock Interpreting black-box models: a review on explainable artificial intelligence.
\newblock \emph{Cognitive Computation}, 16\penalty0 (1):\penalty0 45--74, 2024.

\bibitem[Havasi et~al.(2022)Havasi, Parbhoo, and Doshi-Velez]{havasi2022addressing}
Havasi, M., Parbhoo, S., and Doshi-Velez, F.
\newblock {Addressing leakage in concept bottleneck models}.
\newblock \emph{Advances in Neural Information Processing Systems}, 35:\penalty0 23386--23397, 2022.

\bibitem[He et~al.(2016)He, Zhang, Ren, and Sun]{he2016deep}
He, K., Zhang, X., Ren, S., and Sun, J.
\newblock Deep residual learning for image recognition.
\newblock In \emph{Proceedings of the IEEE Conference on Computer Vision and Pattern Recognition}, pp.\  770--778, 2016.

\bibitem[Kim et~al.(2018)Kim, Wattenberg, Gilmer, Cai, Wexler, Viegas, et~al.]{kim2018interpretability}
Kim, B., Wattenberg, M., Gilmer, J., Cai, C., Wexler, J., Viegas, F., et~al.
\newblock Interpretability beyond feature attribution: Quantitative testing with concept activation vectors (tcav).
\newblock In \emph{International Conference on Machine Learning}, pp.\  2668--2677. PMLR, 2018.

\bibitem[Koh et~al.(2020)Koh, Nguyen, Tang, Mussmann, Pierson, Kim, and Liang]{koh2020concept}
Koh, P.~W., Nguyen, T., Tang, Y.~S., Mussmann, S., Pierson, E., Kim, B., and Liang, P.
\newblock {Concept Bottleneck Models}.
\newblock In \emph{International Conference on Machine Learning}, pp.\  5338--5348. PMLR, 2020.

\bibitem[Oikarinen et~al.(2023)Oikarinen, Das, Nguyen, and Weng]{oikarinen2023labelfree}
Oikarinen, T., Das, S., Nguyen, L.~M., and Weng, T.-W.
\newblock {Label-free Concept Bottleneck Models}.
\newblock In \emph{The Eleventh International Conference on Learning Representations}, 2023.

\bibitem[Ramaswamy et~al.(2020)]{ramaswamy2020ablation}
Ramaswamy, H.~G. et~al.
\newblock Ablation-cam: Visual explanations for deep convolutional network via gradient-free localization.
\newblock In \emph{Proceedings of the IEEE/CVF Winter Conference on Applications of Computer Vision}, pp.\  983--991, 2020.

\bibitem[Selvaraju et~al.(2017)Selvaraju, Cogswell, Das, Vedantam, Parikh, and Batra]{Selvaraju_2017_ICCV}
Selvaraju, R.~R., Cogswell, M., Das, A., Vedantam, R., Parikh, D., and Batra, D.
\newblock {Grad-CAM: Visual Explanations From Deep Networks via Gradient-Based Localization}.
\newblock In \emph{Proceedings of the IEEE International Conference on Computer Vision (ICCV)}, Oct 2017.

\bibitem[Shin et~al.(2023)Shin, Jo, Ahn, and Lee]{shin2023closer}
Shin, S., Jo, Y., Ahn, S., and Lee, N.
\newblock A closer look at the intervention procedure of concept bottleneck models.
\newblock In \emph{International Conference on Machine Learning}, pp.\  31504--31520. PMLR, 2023.

\bibitem[Simonyan \& Zisserman(2014)Simonyan and Zisserman]{simonyan2014very}
Simonyan, K. and Zisserman, A.
\newblock Very deep convolutional networks for large-scale image recognition.
\newblock \emph{arXiv preprint arXiv:1409.1556}, 2014.

\bibitem[Steinmann et~al.(2024)Steinmann, Stammer, Friedrich, and Kersting]{pmlr-v235-steinmann24a}
Steinmann, D., Stammer, W., Friedrich, F., and Kersting, K.
\newblock {Learning to Intervene on Concept Bottlenecks}.
\newblock In Salakhutdinov, R., Kolter, Z., Heller, K., Weller, A., Oliver, N., Scarlett, J., and Berkenkamp, F. (eds.), \emph{Proceedings of the 41st International Conference on Machine Learning}, volume 235 of \emph{Proceedings of Machine Learning Research}, pp.\  46556--46571. PMLR, 21--27 Jul 2024.
\newblock URL \url{https://proceedings.mlr.press/v235/steinmann24a.html}.

\bibitem[Van~der Velden et~al.(2022)Van~der Velden, Kuijf, Gilhuijs, and Viergever]{van2022explainable}
Van~der Velden, B.~H., Kuijf, H.~J., Gilhuijs, K.~G., and Viergever, M.~A.
\newblock Explainable artificial intelligence (xai) in deep learning-based medical image analysis.
\newblock \emph{Medical Image Analysis}, 79:\penalty0 102470, 2022.

\bibitem[Vandenhirtz et~al.(2024)Vandenhirtz, Laguna, Marcinkevi{\v{c}}s, and Vogt]{vandenhirtz2024stochastic}
Vandenhirtz, M., Laguna, S., Marcinkevi{\v{c}}s, R., and Vogt, J.~E.
\newblock {Stochastic Concept Bottleneck Models}.
\newblock In \emph{ICML 2024 Workshop on Structured Probabilistic Inference {\&} Generative Modeling}, 2024.
\newblock URL \url{https://openreview.net/forum?id=8jG3Y0xX7b}.

\bibitem[Wah et~al.(2011)Wah, Branson, Welinder, Perona, and Belongie]{wah2011caltech}
Wah, C., Branson, S., Welinder, P., Perona, P., and Belongie, S.
\newblock The caltech-ucsd birds-200-2011 dataset.
\newblock 2011.

\bibitem[Wang et~al.(2020)Wang, Wang, Du, Yang, Zhang, Ding, Mardziel, and Hu]{wang2020score}
Wang, H., Wang, Z., Du, M., Yang, F., Zhang, Z., Ding, S., Mardziel, P., and Hu, X.
\newblock {Score-CAM: Score-weighted Visual Explanations for Convolutional Neural Networks}.
\newblock In \emph{Proceedings of the IEEE/CVF Conference on Computer Vision and Pattern Recognition Workshops}, pp.\  24--25, 2020.

\bibitem[Yuksekgonul et~al.(2023)Yuksekgonul, Wang, and Zou]{yuksekgonul2023posthoc}
Yuksekgonul, M., Wang, M., and Zou, J.
\newblock {Post-hoc Concept Bottleneck Models}.
\newblock In \emph{The Eleventh International Conference on Learning Representations}, 2023.

\bibitem[Zhou et~al.(2016)Zhou, Khosla, Lapedriza, Oliva, and Torralba]{zhou2016learning}
Zhou, B., Khosla, A., Lapedriza, A., Oliva, A., and Torralba, A.
\newblock {Learning deep features for discriminative localization}.
\newblock In \emph{Proceedings of the IEEE Conference on Computer Vision and Pattern Recognition}, pp.\  2921--2929, 2016.

\end{thebibliography}
\end{document}